\documentclass[journal]{IEEEtran}

\usepackage{amsmath,amsfonts,bm}

\def\1{\bm{1}}

\DeclareMathAlphabet{\mathsfit}{\encodingdefault}{\sfdefault}{m}{sl}
\SetMathAlphabet{\mathsfit}{bold}{\encodingdefault}{\sfdefault}{bx}{n}

\DeclareMathOperator*{\argmin}{arg\,min}

\usepackage[colorlinks=true,citecolor = blue]{hyperref}%
\usepackage{url}            
\usepackage{booktabs}       
\usepackage{amsfonts}       
\usepackage{microtype}      
\allowdisplaybreaks
\usepackage{setspace} 
\usepackage{cite,calc}
\usepackage{algorithmic}
\usepackage{graphicx}
\usepackage{amssymb}

\usepackage{xcolor}
\usepackage{enumerate}
\usepackage{graphicx}
\usepackage{textcomp}
\usepackage{epstopdf}
\usepackage{makecell}

\usepackage{fancyhdr}

\newtheorem{theorem}{Theorem}
\newtheorem{lemma}{Lemma}

\newtheorem{proposition}{Proposition}

\newtheorem{corollary}{Corollary}

\newtheorem{definition}{Definition}
\newtheorem{example}{Example}

\newtheorem{remark}{Remark}

\newcommand{\nn}{\notag}

\DeclareMathOperator*{\gen}{gen}
\newcommand{\genb}{\overline\gen}

\newcommand{\blue}{\color{black}}

\newcommand{\mbE}{\mathbb{E}}
\newcommand{\KLr}{\mathrm{KL}}
\newcommand{\mrR}{\mathrm{R}}

\newcommand{\JDS}{\mathrm{JS}}
\newcommand{\ren}{\mathrm{R}_{\alpha}}

\usepackage{xspace}
\usepackage[colorinlistoftodos, color=blue!30!white 
]{todonotes}                                        
\newcommand{\indep}{\perp \!\!\! \perp}

\newcommand{\mrd}{\mathrm{d}}

\usepackage{xcolor}
\def\BibTeX{{\rm B\kern-.05em{\sc i\kern-.025em b}\kern-.08em
    T\kern-.1667em\lower.7ex\hbox{E}\kern-.125emX}}
\begin{document}

\title{Learning Algorithm Generalization Error Bounds via Auxiliary Distributions}


\author{Gholamali~Aminian\textsuperscript{$\star$},~\IEEEmembership{Member,~IEEE,} Saeed~Masiha\textsuperscript{$\star$},~\IEEEmembership{Member,~IEEE,} 
        Laura~Toni,~\IEEEmembership{Senior~Member,~IEEE,}  
        and~Miguel~R.~D.~Rodrigues,~\IEEEmembership{Fellow,~IEEE}

\thanks{$^*$ Equal Contribution.}
\thanks{One of the ideas of this work was presented, in part, at 2020 IEEE Information Theory Workshop (ITW),  \cite{aminian2021Jensen}.
}
\thanks{G. Aminian is with the Alan Turing Institute, London NW1 2DB, U.K.  (e-mail: gaminian@turing.ac.uk), L. Toni and M. R. D. Rodrigues are with the Electronic and Electrical Engineering Department, University College London, London WC1E 6BT, U.K. (e-mail: \{l.toni, m.rodrigues\}@ucl.ac.uk), and S. Masiha is with the École Polytechnique Fédérale de Lausanne, Lausanne 1015, Switzerland. (e-mail: mohammadsaeed.masiha@epfl.ch).}
}

\maketitle

\begin{abstract}
Generalization error bounds are essential for comprehending how well machine learning models work. In this work, we suggest a novel method, i.e., the Auxiliary Distribution Method, that leads to new upper bounds on expected generalization errors that are appropriate for supervised learning scenarios. We show that our general upper bounds can be specialized under some conditions to new bounds involving the  $\alpha$-Jensen-Shannon, $\alpha$-R\'enyi ($0< \alpha < 1$) information between a random variable modeling the set of training samples and another random variable modeling the set of hypotheses. Our upper bounds based on  $\alpha$-Jensen-Shannon information are also finite. Additionally, we demonstrate how our auxiliary distribution method can be used to derive the upper bounds on excess risk of some learning algorithms in the supervised learning context {\blue and the generalization error under the distribution mismatch scenario in supervised learning algorithms, where the distribution mismatch is modeled as $\alpha$-Jensen-Shannon or $\alpha$-R\'enyi divergence between the distribution of test and training data samples distributions.} We also outline the conditions for which our proposed upper bounds might be tighter than other earlier upper bounds.
\end{abstract}
\begin{IEEEkeywords}
Expected Generalization Error Bounds, population risk upper bound, Mutual Information, $\alpha$-Jensen-Shannon Information, $\alpha$-R\'enyi Information, Distribution mismatch.\end{IEEEkeywords}
\section{Introduction}\label{Sec:Introduction}

\IEEEPARstart{N}{umerous} methods have been proposed in order to describe the generalization error of learning algorithms. These include VC-based bounds~\cite{vapnik1999overview}, algorithmic stability-based bounds~\cite{bousquet2002stability}, algorithmic robustness-based bounds~\cite{xu2012robustness}, PAC-Bayesian bounds~\cite{mcallester2003pac}. Nevertheless, for a number of reasons, many of these generalization error bounds are unable to describe how different machine-learning techniques can generalize: some of the bounds depend only on the hypothesis class and not on the learning algorithm; existing bounds do not easily exploit dependencies between different hypotheses; or do not exploit dependencies between the learning algorithm input and output.

More recently, methods that use information-theoretic tools have also been developed to describe the generalization of learning techniques. Such methods frequently incorporate the many components related to the learning problem by expressing the expected generalization error in terms of certain information measurements between the learning algorithm input (the training dataset) and output (the hypothesis).
In particular, building upon pioneering work by Russo and Zou~\cite{russo2019much}, Xu and Raginsky~\cite{xu2017information} have derived expected generalization error bounds involving the mutual information between the training set and the hypothesis. Bu \textit{et al.}~\cite{bu2020tightening} have derived tighter expected generalization error bounds involving the mutual information between each individual sample in the training set and the hypothesis. Meanwhile, bounds using chaining mutual information have been proposed in~\cite{asadi2018chaining,asadi2020chaining}. Other authors have also constructed information-theoretic based expected generalization error bounds based on other information measures such as $\alpha$-R\'enyi divergence for $\alpha>1$, $f$-divergence, and maximal leakage~\cite{esposito2019generalization}. In \cite{Modak2021Renyi}, an upper bound based on $\alpha$-R\'enyi divergence for $0<\alpha<1$ is derived by using the variational representation of $\alpha$-R\'enyi divergence. Bounds based on the Wasserstein distance between the training sample data and the output of a randomized learning algorithm~\cite{lopez2018generalization}, \cite{wang2019information} and Wasserstein distance between distributions of an individual sample data and the output of the learning algorithm is proposed in \cite{galvez2021tighter}, and tighter upper bounds via convexity of Wasserstein distance are proposed in \cite{aminian2022tighter}. Upper bounds based on conditional mutual information and individual sample conditional mutual information are proposed in \cite{steinke2020reasoning} and \cite{zhou2020individually}, respectively. The combination of conditioning and processing techniques can provide tighter expected generalization error upper bounds~\cite{hafez2020conditioning}. An exact characterization of the expected generalization error for the Gibbs algorithm in terms of symmetrized KL information is provided in \cite{aminian2021exact}. \cite{masiha2021learning} provides information-theoretic expected generalization error upper bounds in the presence of training/test data distribution mismatch, using rate-distortion theory. 

 {\blue Generalization error bounds have also been developed to address scenarios where the training data distribution differs from the test data distribution, known as Distribution Mismatch. This scenario -- which also links to out-of-distribution generalization -- has attracted various contributions in recent years, such as \cite{mansour2009domain,wang2018theoretical,wu2020information}. In particular, Masiha et al. \cite{masiha2021learning} provides information-theoretic generalization error upper bounds in the presence of training/test data distribution mismatch, using rate-distortion theory.}

In this work, we propose an auxiliary distribution method (ADM) to characterize the expected generalization error upper bound of supervised learning algorithms in terms of novel information measures. Our new bounds offer two advantages over existing ones: (1) Some of our bounds -- such as the  $\alpha$-$\JDS$ information ones -- are always finite, whereas conventional mutual information ones (e.g., \cite{xu2017information}) may not be; (2) In contrast to mutual information-based bounds, our bounds—such as the $\alpha$-R\'enyi information for $0<\alpha<1$—are finite for some deterministic supervised learning algorithms; (3) We also apply ADM to provide an upper bound on population risk of supervised learning algorithms under a learning algorithm.

In summary, our main contributions are as follows:
\begin{enumerate}

   \item We suggest a novel method, i.e., ADM, that uses auxiliary distributions over the parameter and data sample spaces to obtain upper bounds on the expected generalization error. 
    \item Using ADM, we derive new expected generalization error bounds expressed via  $\alpha$-$\JDS$ divergence, which is known to be finite.
    
    \item Using ADM, we offer an upper bound based on $\alpha$-R\'enyi divergence for $0<\alpha<1$ with the same convergence rate as the mutual information-based upper bound. Furthermore, in contrast to the mutual information-based bounds, the $\alpha$-R\'enyi divergence bounds for $0<\alpha<1$ can be finite when the hypothesis (output of the learning algorithm) is a deterministic function of at least one data sample.
    
    \item Using our upper bounds on expected generalization error, we also provide upper bounds on excess risk of some learning algorithms as solutions to regularized empirical risk minimization by $\alpha$-R\'enyi or $\alpha$-Jensen-Shannon divergences.

    \item {\blue Using ADM, we also provide generalization error upper bound under training and test data distribution mismatch. It turns out that training and test distribution mismatch is captured in our upper bounds via $\alpha$-Jensen-Shannon or $\alpha$-R\'enyi divergences.}
   
\end{enumerate}

It is noteworthy to add that, although the  $\alpha$-$\JDS$ measure does not appear to have been used to characterize the generalization ability of learning algorithms, these information-theoretic quantities as well as $\alpha$-R\'enyi measure for $0<\alpha<1$, have been employed to study some machine learning problems, including the use of
\begin{itemize}
    \item  $\alpha$-$\JDS$ as a loss function under label noise scenario~\cite{englesson2021generalized}, and Jensen-Shannon divergence ( $\alpha$-$\JDS$ divergence for $\alpha=1/2$) in adversarial learning~\cite{goodfellow2014generative} and active learning~\cite{melville2005active}.
    \item $\alpha$-R\'enyi divergence in feature extraction ~\cite{choi2003feature} and image segmentation based on clustering~\cite{coleman1979image}.
\end{itemize}

\section{Problem Formulation}
\subsection{Notations}
In this work, we adopt the following notation in the sequel. Calligraphic letters denote spaces (e.g. $\mathcal{Z}$), Upper-case letters denote random variables (e.g., $Z$), and lower-case letters denote a realization of random variable (e.g. $z$). We denote the distribution of the random variable $Z$ by $P_Z$, the joint distribution of two random variables $(Z_1,Z_2)$ by $P_{Z_1,Z_2}$, and the $\alpha$-convex combination of the joint distribution $P_{Z_1,Z_2}$ and the product of two marginals $P_{Z_1} \otimes P_{Z_2}$, i.e. $\alpha P_{Z_1}\otimes P_{Z_2} +(1-\alpha) P_{Z_1,Z_2}$ for $\alpha\in (0,1)$, by $P_{Z_1,Z_2}^{(\alpha)}$. The set of 
distributions ( measures) over a space $\mathcal{X}$ with is denoted
$\mathcal{P}(\mathcal{X})$. We denote the derivative of a real-valued function $f(x)$ with respect to its argument $x$ by $f^{\prime}( \cdot )$. We also adopt the notion $\log (\cdot)$ for the natural logarithm. The function $f(x)$ is $L_f$-Lipschitz if $|f(x_1)-f(x_2)|\leq L_f \|x_1-x_2\|_2$, where $\|\cdot\|_2$ is $L_2$-norm. Let $\mathcal{N}(a,B)$ denotes the Gaussian distribution over $\mathbb{R}^d$ with mean $a\in\mathbb{R}^d$ and covariance matrix $B\in\mathbb{R}^{d\times d}$.
\subsection{Framework of Statistical Learning}
We analyze a standard supervised learning setting where we wish to learn a hypothesis given a set of input-output examples that can then be used to predict a new output given a new input.

In particular, in order to formalize this setting, we model the input data (also known as features) using a random variable $X \in \mathcal{X}$ where $\mathcal{X}$ is the input space, and we model the output data (also known as predictors or labels) using a random variable $Y \in \mathcal{Y}$ where $\mathcal{Y}$ is the output space. We also model input-output data pairs using a random variable $Z = (X,Y) \in \mathcal{Z} = \mathcal{X} \times \mathcal{Y}$ where $Z$ is drawn from $\mathcal{Z}$ per some unknown distribution $\mu$. We also let $S = \{Z_i\}_{i=1}^n$ be a training set consisting of $n$ input-output data points drawn i.i.d. from $\mathcal{Z}$ according to $\mathcal{\mu}$. 

Our goal is to learn a parameterized function, $f_W:\mathcal{X} \rightarrow \mathcal{Y}$, where the parameters are a random variable $W \in \mathcal{W}\subset \mathbb{R}^d$ and $\mathcal{W}$ is a parameter space. Finally, we represent a learning algorithm via a Markov kernel that maps a given training set $S$ onto parameter $W$ defined on the parameter space $\mathcal{W}$ according to the probability law $P_{W|S}$.

We introduce a (non-negative) loss function $\ell:\mathcal{W} \times \mathcal{Z}  \rightarrow \mathbb{R}^+$ that measures how well a hypothesis (parameterized function) predicts an output given an input. We can define the population risk and the empirical risk associated with a given hypothesis as follows:
\begin{align}
&L_{\mu}(w):= \int_{\mathcal{Z}}\ell(w,z)d\mu(z),\\
&L_E(w,s):=\frac{1}{n}\sum_{i=1}^n \ell(w,z_i),
\end{align}
respectively. We can also define the (expected) generalization error,
\begin{align}\label{eq: gen-expected}
\genb(P_{W|S},\mu)&:=\mbE_{P_{W,S}}[\text{gen}(W,S,\mu)],
\end{align}
where $\text{gen}(w,s,\mu):= L_{\mu}(w)-L_E(w,s)$. This (expected) generalization error quantifies by how much the population risk deviates from the empirical risk. This quantity cannot be computed directly because $\mu$ is unknown, but it can often be (upper) bounded, thereby providing a means to gauge various learning algorithms' performance. We are also interested in excess risk under the learning algorithm $P_{W|S}$,
\begin{align}\label{eq: excess risk}
\mathcal{E}_r(P_{W|S},\mu)&:=\mbE_{P_{W,S}}[L_{\mu}(W)]-\inf_{w\in\mathcal{W}}L_{\mu}(w).
\end{align}
Note that the excess risk can be decomposed as follows,
\begin{equation*}
\begin{split}
    \mathcal{E}_r (P_{W|S},\mu)=\genb(P_{W|S},\mu)+\mbE_{P_{W,S}}[L_{E}(W,S)]-\inf_{w\in\mathcal{W}}L_{\mu}(w),
\end{split}
\end{equation*}
where the first term is expected generalization error and the second is statistical excess risk.

{\blue Furthermore, we analyse a supervised learning scenario under distribution mismatch ( a.k.a. out-of-distribution), where training and test data are drawn from different distributions ($\mu$ and $\mu'$, respectively).
In particular, we define the population risk based on test distribution $\mu'$ as,
\begin{align}
&L_P(w,\mu')\triangleq \int_{\mathcal{Z}}\ell(w,z)d\mu'(z).
\end{align}
 We define the mismatched(expected) generalization error as
\begin{align}\label{Eqgen}
\genb(P_{W|S},\mu,\mu')&\triangleq\mathbb{E}_{P_{W,S}}[\text{gen}(W,S,\mu,\mu')],
\end{align}
where $\text{gen}(w,s,\mu,\mu')\triangleq L_P(w,\mu')-L_E(w,s)$.}

Our goal in the sequel will be to derive (upper) bounds on the expected generalization errors~\eqref{eq: gen-expected} and the excess risk~\eqref{eq: excess risk} in terms of various information-theoretic measures.

\subsection{Auxiliary Distribution Method}
We describe our main method to derive upper bounds on the expected generalization error, i.e., the ADM. Consider $P$ and $Q$ as two distributions defined on a measurable space $\mathcal{X}$ and let $f:\mathcal{X}\to \mathbb{R}$ be a measurable function. Assume that we can use an asymmetric information measure $T(P\|Q)$ between $P$ and $Q$ to construct the following upper bound:
\begin{align}\label{auxiliary_dist_1001}
   \left| \mbE_{P}[f(X)]-\mbE_{Q}[f(X)]\right|\le F(T(P\|Q)),
\end{align}
where $F(\cdot)$ is a given non-decreasing concave function. 
 
Consider $R$ as an auxiliary distribution on the same space $\mathcal{X}$. We can use the following upper bound instead of \eqref{auxiliary_dist_1001}:
\begin{align}\label{Eq: ADM1}
& \left|\mbE_{P}[f(X)]-\mbE_{Q}[f(X)]\right|\leq\nonumber\\
    &\left|\mbE_{P}[f(X)]-\mbE_{R}[f(X)]\right|+\left|\mbE_{Q}[f(X)]-\mbE_{R}[f(X)]\right|\nonumber\\
&\le F(T(P\|R))+F(T(Q\|R))
\end{align}
 From concavity of $F$, we have
\begin{align}\label{Eq: concave and Pyth}
    &F(T(P\|R))+F(T(Q\|R))\le\nonumber\\
    &2F\bigg(T(P\|R)/2+T(Q\|R)/2\bigg)
\end{align}
We assume that $T$ satisfies a reverse triangle inequality as follows: 
\begin{align}
    \min_{R\in \mathcal{P}(X)} T(P\|R)+T(Q\|R) \le T(P\|Q) .
\end{align}
Considering $R^{*}\in\argmin_{R}T(P\|R)+T(Q\|R)$, we have 
\begin{align}\label{Eq: ADM2}
    &\left|\mbE_{P}[f(X)]-\mbE_{Q}[f(X)]\right|\le\nonumber\\
    &2F\bigg(T(P\|R^{*})/2+T(Q\|R^{*})/2\bigg).
\end{align}
We can also provide another upper bound based on $T(R\|P)$ and $T(R\|Q)$ instead of $T(P\|R)$ and $T(Q\|R)$:
\begin{align}
& \left|\mbE_{P}[f(X)]-\mbE_{Q}[f(X)]\right|\leq\nonumber\\
    &\left|\mbE_{R}[f(X)]-\mbE_{P}[f(X)]\right|+\left|\mbE_{R}[f(X)]-\mbE_{Q}[f(X)]\right|\nonumber\\
&\le F(T(R\|P))+F(T(R\|Q)).
\end{align}
Considering $\tilde{R}\in\argmin_{R\in  \mathcal{P}(X)}T(R\|P)+T(R\|Q)$, we have 
\begin{align}\label{Eq: ADM3}
    &\left|\mbE_{P}[f(X)]-\mbE_{Q}[f(X)]\right|\le\nonumber\\
    &2F\bigg(T(\tilde{R}\|P)/2+T(\tilde{R}\|Q)/2\bigg).
\end{align}
 Via this ADM approach -- taking $T(\cdot\|\cdot)$ to be a KL divergence -- we can derive expected generalization error upper bounds involving KL divergences as follows:
\begin{align}
    \alpha \KLr(P_{W,Z_i}\|\widehat{P}_{W,Z_i}) + (1-\alpha)\KLr(P_W\otimes\mu\|\widehat{P}_{W,Z_i}),\label{conv_KL_1}\\
     \alpha \KLr(\widehat{P}_{W,Z_i}\|P_{W,Z_i}) + (1-\alpha)\KLr(\widehat{P}_{W,Z_i}\|P_W\otimes\mu),\label{conv_KL_2}
\end{align}
where $\widehat{P}_{W,Z_i}$, $P_{W,Z_i}$ and $P_W\otimes\mu$ are an auxiliary joint distribution over the space $\mathcal{Z}\times \mathcal{W}$, the true joint distribution of the random variables $W$ and $Z_i$ and the product of marginal distributions of random variables $W$ and $Z_i$, respectively.
Inspired by the ADM, we use the fact that KL divergence is asymmetric and satisfies the reverse triangle inequality \cite{topsoe2007information}. Hence, we can choose the auxiliary joint distribution, $\widehat{P}_{W,Z_i}$, to derive new upper bounds which are finite or tighter under some conditions.

\subsection{Information Measures} 
In our characterization of the expected generalization error upper bounds, we will use the information measures between two distributions $P_X$ and $P_{X^\prime}$ on a common measurable space $\mathcal{X}$, summarized in Table \ref{table: divergence Measures}. The last two divergences are $\alpha$-$\JDS$ divergence\footnote{a.k.a. capacitory discrimination ~\cite{topsoe2000some} for $\alpha=1/2$}, $\alpha$-R\'enyi divergence, which can be characterized by \eqref{conv_KL_1} and \eqref{conv_KL_2}, respectively (See their characterizations as a convex combination of KL-divergences in Lemmas \ref{lemma: Jensen-Shannon equality} and \ref{lemma: alpha renyi equality}). They are the main divergences discussed in this paper and defined in Table \ref{table: divergence Measures}. KL divergence, Symmetrized KL divergence, Bhattacharyya distance, and Jensen-Shannon divergence can be obtained as special cases of the first three divergences in Table \ref{table: divergence Measures}. 

\begin{table}[!h]
\caption{Divergence Measures Definitions}
\centering
\resizebox{\columnwidth}{!}{\begin{tabular}[h!]{cc}

 \toprule
 Divergence Measure &  Definition\\
 \midrule
 
 KL divergence~\cite{cover1999elements}   
 &
 $\KLr(P_X\|P_{X^\prime}):=\int_\mathcal{X}P_X(x)\log\left(\frac{P_X(x)}{P_{X^\prime}(x)}\right)dx$\\

 \cmidrule(lr{0.5em}){1-2}
   $\alpha$-$\JDS$ divergence~\cite{nielsen2020generalization,lin1991divergence}
 &
 \makecell{$\JDS_{\alpha}(P_{X^\prime}\|P_X):=$ \\ $\alpha \KLr\left(P_X\|\alpha P_X + (1-\alpha) P_{X^\prime}\right)+(1-\alpha)\KLr\left(P_{X^\prime}\|\alpha P_X + (1-\alpha) P_{X^\prime}\right)$}\\
 
\cmidrule(lr{0.5em}){1-2}
Jensen-Shannon divergence~\cite{lin1991divergence}
 &
 \makecell{$\mathrm{JSD}(P_{X^\prime}\|P_X):=\JDS_{1/2}(P_{X^\prime}\|P_X)$ \\ $=\frac{1}{2} \KLr\left(P_X\Big|\Big|\frac{P_X+P_{X^\prime}}{2}\right)+\frac{1}{2}\KLr\left(P_{X^\prime}\Big|\Big|\frac{P_X+P_{X^\prime}}{2}\right)$}\\
 \cmidrule(lr{0.5em}){1-2}
 $\alpha$-R\'enyi divergence for $\alpha \in [0,\infty)$~\cite{van2014renyi}
 &  
 $\ren(P_{X^\prime}\|P_X):=\frac{1}{\alpha-1}\log\left(\int_\mathcal{X}P_X^\alpha(x) P_{X^\prime}^{1-\alpha}(x)dx\right) $  \\
\cmidrule(lr{0.5em}){1-2}
 Bhattacharyya distance~\cite{kailath1967divergence}
 &
 \makecell{$D_{B}(P_{X^\prime}\|P_{X}):= \mathrm{R}_{1/2}(P_{X^\prime}\|P_{X})$\\$=-\log\left(\int_\mathcal{X}\sqrt{P_X(x) P_{X^\prime}(x)}dx\right)$}\\
  \bottomrule
\end{tabular}}

\label{table: divergence Measures}
\end{table}

In addition, in our expected generalization error characterizations, we will also use various information measures between two random variables $X$ and $X'$ with joint distribution $P_{XX'}$ and marginals $P_X$ and $P_{X'}$. These information measures are summarized in Table~\ref{table: information Measures}. Note that all these information measures are zero if and only if the random variables $X$ and $X^\prime$ are independent.

\begin{table}[!h]
\caption{Information Measures Definitions}
\centering
\resizebox{\columnwidth}{!}{\begin{tabular}[h!]{cc}

 \toprule
 Information Measure & Definition\\
 \midrule
 Mutual information 
 &
 $I(X;X^\prime):= \KLr(P_{X,X^\prime}\|P_X\otimes P_{X^\prime})$\\
\cmidrule(lr{0.5em}){1-2}
 Lautum information~\cite{palomar2008lautum}
 &  
 $L(X;X^\prime):= \KLr(P_X\otimes P_{X^\prime}\|P_{X,X^\prime}) $  \\
\cmidrule(lr{0.5em}){1-2}
  $\alpha$-$\JDS$ information ($0<\alpha<1$)
 &
 $I_{\JDS}^\alpha(X;X^\prime):= \JDS_{\alpha}(P_{X,X^\prime}\|P_X\otimes P_{X^\prime})$\\
\cmidrule(lr{0.5em}){1-2}
  Jensen-Shannon information~\cite{sason2016f}
 &
 $I_{\JDS}(X;X^\prime):= \mathrm{JSD}(P_{X,X^\prime}\|P_X\otimes P_{X^\prime})$\\
\cmidrule(lr{0.5em}){1-2}
 $\alpha$-R\'enyi information
 &
 $I_{\mrR}^\alpha(X;X^\prime):= \ren(P_{X,X^\prime}\|P_X\otimes P_{X^\prime})$\\
\cmidrule(lr{0.5em}){1-2}
  Sibson’s $\alpha$-Mutual information~\cite{verdu2015alpha}
 &
 $I_{\mathrm{S}}^{\alpha}(X;X^\prime):= \min_{Q_{X^\prime}} \ren(P_{X,X^\prime}\|P_X\otimes Q_{X^\prime})$\\
 \bottomrule

\end{tabular}}

\label{table: information Measures}
\end{table}
\subsection{Definitions}
We offer some standard definitions that will guide our analysis in the sequel.
\begin{definition}\label{MGF_def}
The cumulant generating function (CGF) of a random variable $X$ is defined as
\begin{equation}
\Lambda_X(\lambda) := \log \mbE[e^{\lambda(X-\mbE[X])}].
\end{equation}
\end{definition}
Assuming $\Lambda_X(\lambda)$ exists, it can be verified that $\Lambda_X(0)=\Lambda_X'(0)=0$, and that it is convex.

\begin{definition}\label{Legendre dual}
For a convex function $\psi$ defined on the interval $[0,b)$, where $0<b\le \infty$, its Legendre dual $\psi^\star$ is defined as
\begin{equation}
  \psi^\star(x) := \sup_{\lambda \in [0,b)} \big(\lambda x-\psi(\lambda)\big).
\end{equation}
\end{definition}

The following lemma characterizes a useful property of the Legendre dual and its inverse function.
\begin{lemma}\label{lemma:psi_star}\cite[Lemma 2.4]{boucheron2013concentration} 
Assume that $\psi(0)= \psi'(0) = 0$. Then, the Legendre dual $\psi^\star(x)$ of $\psi(x)$ defined above is a non-negative convex and non-decreasing function on $[0,\infty)$ with $\psi^\star(0) = 0$. Moreover, its inverse function $\psi^{\star -1}(y) = \inf\{x\ge 0: \psi^\star(x)\ge y\}$ is concave, and can be written as
\begin{equation}
  \psi^{\star -1}(y) = \inf_{\lambda \in [0,b)} \Big( \frac{y+\psi(\lambda)}{\lambda} \Big),\quad b> 0.
\end{equation}
\end{lemma}
Importantly, using these results, we can characterize the tail behaviour of Sub-Gaussian random variables. A random variable $X$ is  $\sigma$-sub-Gaussian, if $\psi(\lambda)= \frac{\sigma^2\lambda^2}{2}$ is an upper bound on $\Lambda_X(\lambda)$, for $\lambda \in \mathbb{R}$. Then by Lemma~\ref{lemma:psi_star}, \begin{align}\label{Eq: sub-Gaussian}
\psi^{\star -1}(y)=\sqrt{2\sigma^2y}.
\end{align}
The tail behaviour of sub-Exponential and sub-Gamma random variables are introduced in \cite{aminian2021exact}.
\section{Upper Bounds on the Expected Generalization Error via ADM}\label{Sec:Auxiliary distribution Generalization Error}
We provide a series of bounds on the expected generalization error of supervised learning algorithms based on different information measures using the ADM coupled with KL divergence.
\subsection{ \texorpdfstring{$\alpha$}--Jensen-Shannon- based Upper Bound}\label{subsection: Jensen-Shannon}
In the following Theorem, we provide a new expected generalization error upper bound based on KL divergence by applying ADM and using KL divergences terms, $\KLr(P_W\otimes\mu\|\widehat{P}_{W,Z_i})$ and $\KLr(P_{W,Z_i}\|\widehat{P}_{W,Z_i})$. Somme proof details are deferred to Appendix~\ref{app: proof of jsd}.
\begin{theorem}\label{Theorem: General Upper Bounds}
Assume that under an auxiliary joint distribution $\widehat{P}_{W,Z_i}\in\mathcal{P}(\mathcal{W}\times \mathcal{Z})$ -- $\Lambda_{\ell(W,Z_i)}(\lambda)$ exists, it is upper bounded by $\psi_+(\lambda)$ for $\lambda \in [0,b_+)$, $0<b_+<+\infty$, and it is also upper bounded by $\psi_-(-\lambda)$ for $\lambda \in (b_-,0]$, $\forall i=1,\cdots,n$. Also assume that $\psi_+(\lambda)$ and $\psi_-(\lambda)$ are convex functions and $\psi_-(0)=\psi_+(0)=\psi_+^\prime(0)=\psi_-^\prime(0)=0$. Then, it holds that:
\begin{align}\label{Eq: Theorem: General Upper Bounds positive}
    \genb(P_{W|S},\mu)&\leq \frac{1}{n} \sum_{i=1}^n \left(\psi_+^{\star -1}(A_i)+\psi_-^{\star -1}(B_i)\right),\\
    -\genb(P_{W|S},\mu)&\leq \frac{1}{n}  \sum_{i=1}^n \left(\psi_-^{\star -1}(A_i)+\psi_+^{\star -1}(B_i)\right),
\end{align}
where $A_i=\KLr(P_W\otimes\mu\|\widehat{P}_{W,Z_i})$, $B_i=\KLr(P_{W,Z_i}\|\widehat{P}_{W,Z_i})$, $
\psi_-^{\star -1}(x)=\inf_{\lambda \in [0,-b_-)}\frac{x+\psi_-(\lambda)}{\lambda}
$
and
$
\psi_+^{\star -1}(x)=\inf_{\lambda \in [0,b_+)}\frac{x+\psi_+(\lambda)}{\lambda}\,.
$ 
\end{theorem}
\begin{IEEEproof}
The proofs of the bounds to $\genb(P_{W|S},\mu)$ and $-\genb(P_{W|S},\mu)$ are similar. Therefore, we focus on the latter. 

Let us consider the Donsker–Varadhan variational representation of KL divergence between two probability distributions $\alpha$ and $\beta$ on a common space $\Psi$ given by~\cite{dupuis2011weak}:
\begin{align}
    \KLr(\alpha\|\beta)=\sup_{f} \int_\Psi f d\alpha -\log\int_\Psi e^f d\beta,
\end{align}
where $f\in \mathcal{F}=\{f:\Psi\rightarrow\mathbb{R} \text{ s.t.  } \mbE_{\beta}[e^f]< \infty \}$.

Using the Donsker-Varadhan representation to bound $\KLr(P_{W,Z_i}\|\widehat{P}_{W,Z_i})$ for $\lambda \in (b_-,0]$ as follows:
\begin{align}\label{Eq:Theorem 1 GE KL Joint}
    &\KLr(P_{W,Z_i}\|\widehat{P}_{W,Z_i})\geq \\\nn &\mbE_{P_{W,Z_i}}[\lambda \ell(W,Z_i)]-\log \mbE_{\widehat{P}_{W,Z_i}}[e^{\lambda \ell(W,{Z}_i)}]\geq\\\label{Eq: Theorem 1 GE KL CGF bound}
    &\lambda(\mbE_{P_{W,Z_i}}[ \ell(W,Z_i)]- \mbE_{\widehat{P}_{W,Z_i}}[\ell(W,{Z}_i)])-\psi_-(-\lambda),
\end{align}
where the last inequality is due to:
\begin{align}
  &\Lambda_{\ell({W},{Z}_i)}(\lambda)=\\\nn
  &\log\left(\mbE_{\widehat{P}_{W,Z_i}}[e^{\lambda(\ell(W,{Z}_i)-\mbE_{\widehat{P}_{W,Z_i}}[\ell(W,{Z}_i)])}]\right)\leq \psi_-(-\lambda).
\end{align}
 It can then be shown from \eqref{Eq: Theorem 1 GE KL CGF bound} that the following holds for $\lambda \in (b_-,0]$:
\begin{align}\label{Eq: GE upper BU 2}
    &\mbE_{\widehat{P}_{W,Z_i}}[\ell(W,{Z}_i)] - \mbE_{{P}_{W,Z_i}}[ \ell(W,Z_i)]\leq\\\nn
    &\inf_{\lambda \in [0,-b_-)} \frac{\KLr(P_{W,Z_i}\|\widehat{P}_{W,Z_i})+\psi_-(\lambda)}{\lambda}=\\
    &\psi_-^{\star -1}(\KLr(P_{W,Z_i}\|\widehat{P}_{W,Z_i})).
\end{align}

It can likewise also be shown by adopting similar steps that the following holds for $\lambda \in [0,b_+)$:
\begin{align}
    &\mbE_{P_{W,Z_i}}[ \ell(W,Z_i)]-\mbE_{\widehat{P}_{W,Z_i}}[\ell(W,Z_i)]\leq\\\nn
    &\inf_{\lambda \in [0,b_+)} \frac{\KLr(P_{W,Z_i}\|\widehat{P}_{W,Z_i})+\psi(\lambda)}{\lambda}=\\
    &\psi_+^{\star -1}(\KLr(P_{W,Z_i}\|\widehat{P}_{W,Z_i})).
\end{align}

We can similarly show using an identical procedure that:
\begin{align}\label{Eq: GE upper BU 1}
    &\mbE_{P_{W}\otimes \mu}[\ell({W},{Z}_i)]- \mbE_{\widehat{P}_{W,Z_i}}[\ell(W,\widehat{Z}_i)] \nn\\
    &\leq
    \psi_+^{\star -1}(\KLr(P_{W} \otimes \mu\|\widehat{P}_{W,Z_i}))\\
    &\mbE_{\widehat{P}_{W,Z_i}}[\ell(W,\widehat{Z}_i)]-\mbE_{P_{W}\otimes \mu}[ \ell({W},{Z}_i)]\nn\\
    &\leq\psi_-^{\star -1}(\KLr(P_{W} \otimes \mu\|\widehat{P}_{W,Z_i})).
\end{align}

Finally, we can immediately bound the expected generalization error by leveraging \eqref{Eq: GE upper BU 1} and \eqref{Eq: GE upper BU 2} as follows:
\begin{align*}
   &\genb(P_{W|S},\mu)=
   \frac{1}{n}\sum_{i=1}^n\mbE_{P_{W}\otimes \mu}[\ell({W},{Z}_i)]-\mbE_{{P}_{W,Z_i}}[\ell(W,Z_i)]\\
   &=\frac{1}{n}\sum_{i=1}^n \mbE_{P_{W}\otimes \mu}[\ell({W},{Z}_i)]- \mbE_{\widehat{P}_{W,Z_i}}[\ell({W},{Z}_i)]
    +\\
    &~\mbE_{\widehat{P}_{W,Z_i}}[\ell({W},{Z}_i)]-\mbE_{{P}_{W,Z_i}}[ \ell(W,Z_i)]\\
   & \leq\frac{1}{n}\sum_{i=1}^n \left(\psi_+^{\star -1}(A_i)+\psi_-^{\star -1}(B_i)\right),
\end{align*}
where $A_i=\KLr(P_W\otimes\mu\|\widehat{P}_{W,Z_i})$ and $B_i=\KLr(P_{W,Z_i}\|\widehat{P}_{W,Z_i})$.
\end{IEEEproof}

Note that Theorem~\ref{Theorem: General Upper Bounds} can be applied to sub-Gaussian~\eqref{Eq: sub-Gaussian}. It can also sub-Exponential and sub-Gamma assumptions on loss function CGF, introduced in \cite{aminian2021exact}. 

We can utilize Theorem~\ref{Theorem: General Upper Bounds} to recover existing expected generalization error bounds and offer new ones. For example, we can immediately recover the mutual information bound \cite{xu2017information} from the following results.

\begin{example}\label{example: mutual information}
 Choose $\widehat{P}_{W,Z_i} = P_W\otimes\mu$ for $i=1,\cdots,n$. It follows immediately from Theorem \ref{Theorem: General Upper Bounds} that:
       \begin{align}\label{Eq:Corollary: other bounds mutual information}
          \genb(P_{W|S},\mu)&\leq \frac{1}{n} \sum_{i=1}^n \psi_-^{\star -1}(I(W;Z_i)),\\
          -\genb(P_{W|S},\mu)&\leq \frac{1}{n} \sum_{i=1}^n \psi_+^{\star -1}(I(W;Z_i)).
        \end{align} 
\end{example}  
\begin{example}\label{example: Lautum information}
Choose $\widehat{P}_{W,Z_i} = P_{W,Z_i}$ for $i=1,\cdots,n$. It also follows immediately from Theorem \ref{Theorem: General Upper Bounds} that:
         \begin{align}\label{Eq:Corollary: other bounds Lautum information}
          \genb(P_{W|S},\mu)&\leq \frac{1}{n} \sum_{i=1}^n \psi_+^{\star -1}(L(W;Z_i)),\\
          -\genb(P_{W|S},\mu)&\leq \frac{1}{n} \sum_{i=1}^n \psi_-^{\star -1}(L(W;Z_i)).
        \end{align} 
\end{example}
The result in Example~\ref{example: mutual information} is the same as a result appearing in~\cite{bu2020tightening} whereas the result in Example~\ref{example: Lautum information} extends the result appearing in \cite{gastpar2019Lautum}.

The conclusion in Theorem~\ref{Theorem: General Upper Bounds} can be extended to many auxiliary distributions by repeatedly using ADM. In this study, we take into account just one auxiliary distribution and use ADM just once.

Building upon Theorem~\ref{Theorem: General Upper Bounds}, we are also able to provide an expected generalization error upper bound based on a convex combination of KL terms, i.e.,
$$ \alpha \KLr(P_W\otimes\mu\|\widehat{P}_{W,Z_i}) + (1-\alpha) \KLr(P_{W,Z_i}\|\widehat{P}_{W,Z_i}),$$
that relies on a certain $\sigma$-sub-Gaussian tail assumption.
\begin{proposition}\label{Prop: Avergae Direct Upper bound}
Assume that the loss function is $\hat{\sigma}$-sub-Gaussian-- under the distribution $\widehat{P}_{W,Z_i}$ $\forall i=1,\cdots,n$-- Then, it holds $\forall \alpha \in (0,1)$ that:
\begin{align}\label{Eq: Proposition: General Upper Bounds}
    &|\genb(P_{W|S},\mu)|\leq \frac{1}{n}  \sum_{i=1}^n \sqrt{2\hat{\sigma}^2\frac{\left(\alpha A_i+ (1-\alpha)B_i\right)}{\alpha (1-\alpha)}},
\end{align}
where $A_i=\KLr(P_{W,Z_i}\|\widehat{P}_{W,Z_i})$ and $B_i=\KLr(P_W\otimes\mu\|\widehat{P}_{W,Z_i})$\,.
\end{proposition}
\begin{IEEEproof}
The assumption that the loss function is $\sigma$-sub-Gaussian under the distribution $\widehat{P}_{W,Z_i}$ implies that $\psi_-^{\star -1}(y)=\psi_+^{\star -1}(y)=\sqrt{2\sigma^2y}$, \cite{bu2020tightening}. Consider the arbitrary auxiliary distributions $\{\widehat{P}_{W,Z_i}\}_{i=1}^n$ defined on $\mathcal{W}\times \mathcal{Z}$.
\begin{align}\label{Eq: Ex 31}
    &\overline{\text{\rm{gen}}}(\mu,P_{W|S})=\mbE_{P_{W}P_{S}}[L_E(W,S)]-\mbE_{P_{W,S}}[L_E(W,S)]\nn\\
    &=\frac{1}{n}\sum_{i=1}^n \mbE_{P_{W}P_{Z_i}}[\ell(W,Z_i)]- \mbE_{P_{W Z_i}}[\ell(W,Z_i)] \\
    &\leq \frac{1}{n}\sum_{i=1}^n \left|\mbE_{P_{W}P_{Z_i}}[\ell(W,Z_i)]- \mbE_{P_{W Z_i}}[\ell(W,Z_i)]\right|
\end{align}

Using the assumption that the loss function $\ell(w,z_i)$ is $\hat{\sigma}^2$-sub-Gaussian under distribution $\widehat{P}_{W,Z_i}$ and Donsker-Varadhan representation for $\KLr(P_{W Z_i}\|\widehat{P}_{W,Z_i})$, we have:
\begin{align}\label{1001}
&\lambda\left(\mbE_{P_{W ,Z_i}}[\ell(W,Z_i)]-\mbE_{\widehat{P}_{W,Z_i}}[\ell(W,Z_i)]\right) \le\\\nn&\quad \KLr(P_{W Z_i}\|\widehat{P}_{W,Z_i})+\frac{\lambda^2\hat{\sigma}^2}{2}.\quad\forall\lambda\in \mathbb{R}
\end{align}

Using the assumption loss that the function $\ell(w,z_i)$ is $\hat{\sigma}^2$-sub-Gaussian under distribution $\widehat{P}_{W,Z_i}$ and Donsker-Varadhan representation for $\KLr(\widehat{P}_{W,Z_i}\|P_{W}P_{Z_i})$, we have,
\begin{align}\label{10021}
& \lambda'\left(\mbE_{P_{W}P_{Z_i}}[\ell(W,Z_i)]-\mbE_{\widehat{P}_{W,Z_i}}[\ell(W,Z_i)]\right)\le\\\nn&\quad \KLr(P_{W}P_{Z_i} \| \widehat{P}_{W,Z_i})+\frac{{\lambda'}^2\hat{\sigma}^2}{2},\quad\forall\lambda'\in \mathbb{R}.
\end{align}
Assuming $\lambda<0$, then we can choose $\lambda'=\frac{\alpha}{\alpha-1} \lambda$. Hence we have,
\begin{align}\label{10010}
&\mbE_{\widehat{P}_{W,Z_i}}[\ell(W,Z_i)] - \mbE_{P_{W ,Z_i}}[\ell(W,Z_i)] \le\\\nn&\quad \frac{\KLr(P_{W,Z_i}\|\widehat{P}_{W,Z_i})}{|\lambda|}+\frac{|\lambda|\hat{\sigma}^2}{2},\quad\forall\lambda\in \mathbb{R^-},
\end{align}
and,
\begin{align}\label{10020}
&\mbE_{P_{W}P_{Z_i}}[\ell(W,Z_i)]-\mbE_{\widehat{P}_{W,Z_i}}[\ell(W,Z_i)]\le \\\nn&\quad\frac{\KLr(P_{W}P_{Z_i}\| \widehat{P}_{W,Z_i})}{\lambda'}+\frac{\lambda'\hat{\sigma}^2}{2},\quad\forall\lambda'\in \mathbb{R^+}.
\end{align}
Merging two Inequalities \eqref{10010} and \eqref{10020}, we have
\begin{align}\label{Eq: Ex 5}
    &\mbE_{P_W P_{Z_i}}[\ell(W,Z_i)]-\mbE_{P_{W Z_i}}[\ell(W,Z_i)]\le\\\nn
    &\frac{\alpha \KLr(P_{W,Z_i}\|\widehat{P}_{W,Z_i})+(1-\alpha)\KLr(P_{W}P_{Z_i}\|\widehat{P}_{W,Z_i})}{\alpha|\lambda|}+\\\nn&\quad\frac{|\lambda|\hat{\sigma}^2}{2}+\frac{|\lambda|\frac{\alpha}{1-\alpha}\hat{\sigma}^2}{2},\quad\forall\lambda\in\mathbb{R^-}.
\end{align}
Similarly, using an identical approach, we also obtain,
\begin{align}\label{Eq: Ex 6}
    &-\left(\mbE_{P_W P_{Z_i}}[\ell(W,Z_i)]-\mbE_{P_{W Z_i}}[\ell(W,Z_i)]\right)\le\\\nn
    &\frac{\alpha \KLr(P_{W,Z_i}\|\widehat{P}_{W,Z_i})+(1-\alpha)\KLr(P_{W}P_{Z_i}\|\widehat{P}_{W,Z_i})}{\alpha\lambda}+\\\nn&\quad\frac{\lambda\hat{\sigma}^2}{2}+\frac{\lambda\frac{\alpha}{1-\alpha}\hat{\sigma}^2}{2},\quad\forall\lambda\in\mathbb{R^+}.
\end{align}
Considering \eqref{Eq: Ex 5} and \eqref{Eq: Ex 6}, we have a non-negative parabola in $\lambda$, whose discriminant must be non-positive, and we have $\forall \alpha \in (0,1)$,
\begin{align}
 & \left | \mbE_{P_W P_{Z_i}}[\ell(W,Z_i)]-\mbE_{P_{W Z_i}}[\ell(W,Z_i)]\right|^{2}\le\\\nn& 2\hat{\sigma}^2\frac{\left(\alpha \KLr(P_{W,Z_i}\|\widehat{P}_{W,Z_i})+(1-\alpha)\KLr(P_{W}P_{Z_i}\|\widehat{P}_{W,Z_i})\right)}{\alpha (1-\alpha)}.
\end{align}
Using \eqref{Eq: Ex 31}, completes the proof.
\end{IEEEproof}

  We propose a Lemma connecting certain KL divergences to the  $\alpha$-$\JDS$ information.

\begin{lemma}\label{lemma: Jensen-Shannon equality}
 Consider an auxiliary distribution $\widehat{P}_{W,Z_i}\in\mathcal{P}(\mathcal{W}\times \mathcal{Z})$. Then, the following equality holds:
\begin{align*}
    &\alpha \KLr(P_W\otimes\mu\|\widehat{P}_{W,Z_i})+(1-\alpha)\KLr(P_{W,Z_i}\|\widehat{P}_{W,Z_i})=\\\nn
    & I_{\JDS}^\alpha (W;Z_i)+\KLr(P_{W,Z_i}^{(\alpha)}\|\widehat{P}_{W,Z_i}).
\end{align*}
\end{lemma}
Note that the proof is inspired by \cite{topsoe2002inequalities}. 

Using the result in Proposition~\ref{Prop: Avergae Direct Upper bound} and ADM we can provide a tighter upper bound. For this purpose, Lemma~\ref{lemma: Jensen-Shannon equality} paves the way to apply ADM and offer a tighter version of the expected generalization error bound appearing in Proposition~\ref{Prop: Avergae Direct Upper bound} based on choosing an appropriate auxiliary distribution, as well as recover existing ones.
\begin{theorem}\label{thm: Minimum general js upper Bound}
Assume that the loss function is $\sigma_{(\alpha)}$-sub-Gaussian-- under the distribution $P_{W,Z_i}^{(\alpha)}$ $\forall i=1,\cdots,n$-- Then, it holds $\forall \alpha \in (0,1)$ that:
\begin{equation*}
    |\genb(P_{W|S},\mu)|\leq \frac{1}{n}  \sum_{i=1}^n \sqrt{2\sigma_{(\alpha)}^2\frac{I_{\JDS}^\alpha (W;Z_i)}{\alpha (1-\alpha)}}, \quad \forall \alpha \in (0,1).
\end{equation*}
\end{theorem}

 The bound in Theorem~\ref{thm: Minimum general js upper Bound} results from minimizing the term $\alpha \KLr(P_W\otimes\mu\|\widehat{P}_{W,Z_i})+(1-\alpha)\KLr(P_{W,Z_i}\|\widehat{P}_{W,Z_i}),$ in the upper bound \eqref{Eq: Proposition: General Upper Bounds}, presented in Proposition~\ref{Prop: Avergae Direct Upper bound}, over the joint auxiliary distribution $\widehat{P}_{W,Z_i}$. Such an optimal joint auxiliary distribution is $P_{W,Z_i}^{(\alpha)}:=\alpha P_{W}P_{Z_{i}}+(1-\alpha)P_{W,Z_{i}}$. Note that, the parameter of sub-Gaussianity, denoted as $\hat{\sigma}$ in Proposition~\ref{Prop: Avergae Direct Upper bound}, relies on $\widehat{P}_{W,Z_i}$. Consequently, the upper bound mentioned in Theorem~\ref{thm: Minimum general js upper Bound} is not the minimum of the upper bound presented in Proposition~\ref{Prop: Avergae Direct Upper bound}. However, assuming a bounded loss function, the upper bound in Theorem~\ref{thm: Minimum general js upper Bound} becomes the minimum of the upper bound in Proposition~\ref{Prop: Avergae Direct Upper bound}.

It turns out that we can immediately recover existing bounds from Theorem~\ref{thm: Minimum general js upper Bound} depending on how we choose $\alpha$.
\begin{remark}[Recovering upper bound based on Jensen-Shannon information]
The expected generalization error upper bound based on Jensen-Shannon information in \cite{aminian2021Jensen} can be immediately recovered by considering $\alpha=\frac{1}{2}$ in Theorem~\ref{thm: Minimum general js upper Bound}.
\end{remark}
\begin{remark}[Recovering upper bounds based on mutual information and lautum information]
The expected generalization error upper bound based on mutual information in \cite{bu2020tightening} and lautum information in \cite{gastpar2019Lautum} can be immediately recovered by considering $\alpha \rightarrow 1$ and $\alpha \rightarrow 0$ in Theorem~\ref{thm: Minimum general js upper Bound}, respectively.
\end{remark}
Note that we can also establish how the bound in Theorem~\ref{thm: Minimum general js upper Bound} behaves as a function of the number of training samples. This can be done by using $\widehat{P}_{W,Z_i}=P_W\otimes\mu$ in Lemma~\ref{lemma: Jensen-Shannon equality}, leading up to
\begin{align}\label{Eq: genaral jensen inequality}
    (1-\alpha)I(W;Z_i)= I_{\JDS}^\alpha (W;Z_i)+ \KLr(P_{W,Z_i}^{(\alpha)}\|P_W\otimes\mu).
\end{align}
and in turn to the following inequality
\begin{align}\label{Eq: main JS and MI inequality }
    I_{\JDS}^\alpha (W;Z_i)\leq (1-\alpha)I(W;Z_i),\quad \forall \alpha\in (0,1).
\end{align}
We prove the convergence rate of the upper bound in Theorem~\ref{thm: Minimum general js upper Bound} using \eqref{Eq: main JS and MI inequality }.
\begin{proposition}\label{Proposition: decay rate Jensen-Shannon}
Assume the hypothesis space is finite and the data samples, $\{ Z_i\}_{i=1}^n$, are i.i.d. Then, the bound in Theorem~\ref{thm: Minimum general js upper Bound} has a convergence rate of $\mathcal{O}(\frac{1}{\sqrt{n}})$. 
\end{proposition}

The value of this new proposed bound presented in Theorem~\ref{thm: Minimum general js upper Bound} in relation to existing bounds can also be further appreciated by offering two additional results.

\begin{proposition}\label{proposition: Bounded general JS upper}  Consider the assumptions in Theorem~\ref{thm: Minimum general js upper Bound}. Then, it follows that:
\begin{align}\label{Eq: proposition: Bounded general JS upper}
    |\genb(P_{W|S},\mu)|\leq \sigma_{(\alpha)} \sqrt{2\frac{h(\alpha)}{\alpha (1-\alpha)}}, \quad \forall \alpha\in (0,1),
\end{align}
where $h(\alpha)=-\alpha\log(\alpha)-(1-\alpha)\log(1-\alpha)$.
\end{proposition}
This proposition shows that, unlike the mutual information-based and lautum information-based generalization bounds that currently exist (e.g. \cite{xu2017information}, \cite{bu2020tightening}, \cite{asadi2018chaining}, and \cite{esposito2019generalization}) the proposed  $\alpha$-$\JDS$ information generalization bound is always finite. {\blue We can also optimize the bound in \eqref{Eq: proposition: Bounded general JS upper} with respect to $\alpha$, where the minimum is achieved at $\alpha=1/2$.}

\begin{corollary}\label{corollary: const upper bound optimum}
 Consider the assumptions in Theorem~\ref{thm: Minimum general js upper Bound}. Then, it follows that:
\begin{align}\label{Eq: proposition: Bounded JS upper}
    |\genb(P_{W|S},\mu)|\leq 2\sigma_{(1/2)} \sqrt{2\log(2)}.
\end{align}
\end{corollary}
 Also, this result applies independently of whether the loss function is bounded or not. Naturally, it is possible to show that the absolute value of the expected generalization error is always upper bounded as follows $|\genb(P_{W|S},\mu)|\leq (b-a)$ for any bounded loss function within the interval $[a,b]$. If we consider the bounded loss functions in the interval $[a,b]$, then our upper bound~\eqref{Eq: proposition: Bounded JS upper} would be $\sqrt{2\log(2)}(b-a)$ which is less than total variation constant upper bound, $2(b-a)$ presented in \cite{raginsky2016information,galvez2021tighter}.

 It is worthwhile to mention that our result cannot be immediately recovered from existing approaches such as~\cite[Theorem.~2]{esposito2019generalization}. For example, if we consider the upper bound based on Jensen-Shannon information, then there exist $f$-divergence based representations of the Jensen-Shannon information as follows:
\begin{align}
    \mathrm{JSD}(P_X,P_{X^\prime})=\int \mrd P_{X} f\left(\frac{\mrd P_{X^\prime}}{\mrd P_X}\right),
\end{align}
with $f(t)=t\log(t)-(1+t)\log(\frac{1+t}{2})$. However, \cite[Theorem.~2]{esposito2019generalization} requires that the function $f(t)$ associated with the $f$-divergence is non-decreasing within the interval $[0,+\infty)$, but such a requirement is naturally violated by the function $f(t)=t\log(t)-(1+t)\log(\frac{1+t}{2})$ associated with the Jensen-Shannon divergence. 

\subsection{\texorpdfstring{$\alpha$}--R\'enyi-based Upper Bound}\label{subsection: Renyi bounds}
Next, we provide a new expected generalization error upper bound based on KL divergence by applying ADM and using the following KL divergences terms, $\KLr(\widehat{P}_{W,Z_i}\|P_W\otimes\mu)$ and $\KLr(\widehat{P}_{W,Z_i}\| P_{W,Z_i})$. All the proof details are deferred to Appendix~\ref{app: proof of renyi}.

\begin{proposition}\label{Prop: Reverse General upper bound}
Suppose that $\Lambda_{\ell(W,Z)}(\lambda) \leq \gamma_+(\lambda)$ and $\Lambda_{\ell(W,Z_i)}(\lambda) \leq \phi_+(\lambda), \quad i=1,\cdots,n$ for $\lambda \in [0,a_+)$, $0<a_+<+\infty$ and $\lambda \in [0,c_+)$, $0<c_+<+\infty$, under $P_W\otimes\mu$ and $P_{W,Z_i}$, resp. We also have $\Lambda_{\ell(W,Z)}(\lambda) \leq \gamma_-(-\lambda)$ and $\Lambda_{\ell(\widetilde{W},\widetilde{Z}_i)}(\lambda) \leq \phi_-(-\lambda), \quad i=1,\cdots,n$ for $\lambda \in (a_-,0]$, $-\infty<a_-<0$ and $\lambda \in (c_-,0]$, $-\infty<c_-<0$ under $P_W\otimes\mu$ and $P_{W,Z_i}$, resp. Assume that $\gamma_+(\lambda)$, $\phi_+(\lambda)$, $\gamma_-(\lambda)$ and $\phi_-(\lambda)$ are convex functions, $\gamma_-(0)=\gamma_+(0)=\gamma_+^\prime(0)=\gamma_-^\prime(0)=0$ and $\phi_-(0)=\phi_+(0)=\phi_+^\prime(0)=\phi_-^\prime(0)=0$. Then, the following upper bounds hold,
\begin{align}\label{Eq: proposition: General Upper Bounds second}
    &\genb(P_{W|S},\mu)\leq \frac{1}{n}  \sum_{i=1}^n \left(\gamma_-^{\star -1}(D_i)+\phi_+^{\star -1}(C_i)\right),\\
    &-\genb(P_{W|S},\mu)\leq \frac{1}{n} \sum_{i=1}^n \left(\phi_-^{\star -1}(C_i)+\gamma_+^{\star -1}(D_i)\right),
\end{align}
where $D_i=\KLr(\widehat{P}_{W,Z_i}\|P_W\otimes\mu)$, $C_i=\KLr(\widehat{P}_{W,Z_i}\|P_{W,Z_i})$, $\gamma_-^{\star -1}(x)=\inf_{\lambda \in [0,-a_-)}\frac{x+\gamma_-(\lambda)}{\lambda}$, \newline$\gamma_+^{\star -1}(x)=\inf_{\lambda \in [0,a_+)}\frac{x+\gamma_+(\lambda)}{\lambda}$, $\phi_-^{\star -1}(x)=\inf_{\lambda \in [0,-c_-)}\frac{x+\phi_-(\lambda)}{\lambda}$ and $\phi_+^{\star -1}(x)=\inf_{\lambda \in [0,c_+)}\frac{x+\phi_+(\lambda)}{\lambda}$.
\end{proposition}
\begin{IEEEproof}
The proof approach is similar to Theorem~\ref{Theorem: General Upper Bounds} by considering different cumulant generating functions and their upper bounds.
\end{IEEEproof}

Inspired by the upper bound in Proposition~\ref{Prop: Reverse General upper bound},
we can provide an upper bound on expected generalization error instantly that is dependent on the convex combination of KL divergence terms, i.e.,  
$$ \alpha \KLr(\widehat{P}_{W,Z_i}\|P_{W,Z_i}) + (1-\alpha)\KLr(\widehat{P}_{W,Z_i}\|P_W\otimes\mu),$$
and assuming $\sigma$-sub-Gaussian tail distribution.
\begin{proposition}[Upper bound with Sub-Gaussian assumption]\label{Prop: reverse upper bound sigma subgaussian} Assume that the loss function is $\sigma$-sub-Gaussian under distribution $P_W \otimes \mu$ and $\gamma$-sub-Gaussian under $P_{W,Z_i}$ $\forall i=1,\cdots,n$-- Then, it holds for $\forall \alpha \in (0,1)$ that,
\begin{align}\label{Eq: Theorem: reverse upper bound }
    &|\genb(P_{W|S},\mu)|\leq \\\nn&\quad \frac{1}{n}  \sum_{i=1}^n \sqrt{2(\alpha\sigma^2+(1-\alpha)\gamma^2)\frac{\left(\alpha C_i+(1-\alpha)D_i\right)}{\alpha (1-\alpha)}},
\end{align}
where $C_i=\KLr(\widehat{P}_{W,Z_i}\| P_W\otimes\mu)$ and $D_i=\KLr(\widehat{P}_{W,Z_i} \| P_{W,Z_i})$\,.
\end{proposition}
Akin to Proposition~\ref{Prop: Avergae Direct Upper bound}, the result in Proposition~\ref{Prop: reverse upper bound sigma subgaussian} paves the way to offer new tighter expected generalization error upper bound by ADM. We next offer a Lemma connecting certain KL divergences to the $\alpha$-R\'enyi information~\cite[Theorem~30]{van2014renyi}.

\begin{lemma}\label{lemma: alpha renyi equality} Consider an arbitrary distribution $\widehat{P}_{W,Z_i}$. Then, the following equality holds for $\forall \alpha \in (0,1)$,
\begin{align}
    &\alpha \KLr(\widehat{P}_{W,Z_i} \| P_W\otimes\mu)+(1-\alpha)\KLr(\widehat{P}_{W,Z_i} \| P_{W,Z_i})=\\\nn
    &\quad (1-\alpha) I_{\mrR}^\alpha(W;Z_i)\\\nn&\quad+\KLr\left(\widehat{P}_{W,Z_i}\|\frac{(P_{Z_i}\otimes P_W)^\alpha (P_{W,Z_i})^{(1-\alpha)}}{\int_{\mathcal{W}\times \mathcal{Z}}(\mrd P_{Z_i}\otimes \mrd P_W)^\alpha (\mrd P_{W,Z_i})^{(1-\alpha)}}\right).
\end{align}
\end{lemma}

A tighter version of the expected generalization error bound appears in Proposition~\ref{Prop: reverse upper bound sigma subgaussian} via ADM and using Lemma~\ref{lemma: alpha renyi equality}.

\begin{theorem}[Upper bound based on $\alpha$-R\'enyi information]\label{thm: Minimum upper Bound alpha Renyi} Consider the same assumptions in Proposition~\ref{Prop: reverse upper bound sigma subgaussian}.  The following upper bound for $\forall \alpha \in (0,1)$ holds,

\begin{align}\label{Eq: Theorem: reverse upper bound 1}
    &|\genb(P_{W|S},\mu)|\leq \frac{1}{n}  \sum_{i=1}^n \sqrt{2(\alpha\sigma^2+(1-\alpha)\gamma^2)\frac{I_{\mrR}^\alpha(W;Z_i)}{\alpha}}.
\end{align}
\end{theorem}

The bound in Theorem~\ref{thm: Minimum upper Bound alpha Renyi} results from minimizing the bound in Proposition~\ref{Prop: reverse upper bound sigma subgaussian} over the joint auxiliary distribution $\hat{P}_{W,Z_i}\in\mathcal{P}(\mathcal{W}\times \mathcal{Z})$. Such an optimal joint auxiliary distribution is 
$$\widehat{P}_{W,Z_i}=\frac{(P_{Z_i}\otimes P_W)^\alpha (P_{W,Z_i})^{(1-\alpha)}}{\int_{\mathcal{W}\times \mathcal{Z}}(\mrd P_{Z_i}\otimes \mrd P_W)^\alpha (\mrd P_{W,Z_i})^{(1-\alpha)} }.$$

\begin{remark}[Deterministic algorithms per sample]
If the parameter, $W$, is a deterministic function of data sample $Z_i$, then $I(W;Z_i)$ is not well-defined as $P_{W,Z_{i}}$ is not \textit{absolutely continuous}\footnote{We say $\mu\ll\nu$, i.e., $\mu$ is absolutely continuous with respect to $\nu$ if $\nu(A)=0$ for some $A\in\mathcal{X}$, then $\mu(A)=0$.} with respect to $P_{W}P_{Z_{i}}$. However, by considering the $\alpha$-R\'enyi information for $\alpha \in[0,1)$,  we do not need to assume the absolute continuous.
\end{remark}

\begin{remark}[Upper bound based on the Bhattacharyya distance]
We can derive the expected generalization error upper bound based on Bhattacharyya distance by considering $\alpha=1/2$ in Theorem~\ref{thm: Minimum upper Bound alpha Renyi}, 
\begin{align*}
    &|\genb(P_{W|S},\mu)|\leq \frac{2}{n}  \sum_{i=1}^n \sqrt{(\sigma^2+\gamma^2)D_B(P_{W,Z_i}\|P_W\otimes \mu)}, 
\end{align*}
\end{remark}
\begin{remark}[Recovering the upper bound based on mutual information and lautum information]
We can recover the expected generalization error upper bound based on mutual information in \cite{xu2017information} and lautum information in \cite{gastpar2019Lautum} by considering $\alpha \rightarrow 1$ and $\alpha \rightarrow 0$ in Theorem~\ref{thm: Minimum upper Bound alpha Renyi}, respectively.
\end{remark}
By considering $\widehat{P}_{W,Z_i}=P_{W,Z_i}$, we have,
\begin{align}
    &\alpha I(W;Z_i)=(1-\alpha) I_{\mrR}^\alpha(W;Z_i)\\\nn&\quad+\KLr\left(P_{W,Z_i}\|\frac{(P_{Z_i}\otimes P_W)^\alpha (P_{W,Z_i})^{(1-\alpha)}}{\int_{\mathcal{W}\times \mathcal{Z}}(\mrd P_{Z_i}\otimes \mrd P_W)^\alpha (\mrd P_{W,Z_i})^{(1-\alpha)} }\right).
\end{align}
Since that KL divergence is non-negative, based on Lemma~\ref{lemma: alpha renyi equality} and the monotonicity of $\ren$ with respect to $\alpha$, we have,
\begin{align}\label{Eq: Renyi compare to KL}
    I_{\mrR}^\alpha(W;Z_i)\leq \min\left\{1,\frac{\alpha}{1-\alpha}\right\} I(W;Z_i).
\end{align}

The result in \eqref{Eq: Renyi compare to KL} implies that our expected generalization error bound based on $\alpha$-R\'enyi information in Theorem~\ref{thm: Minimum upper Bound alpha Renyi} exhibits the same convergence rate as upper bound based on mutual information~\cite{xu2017information}.
\begin{proposition}[Convergence rate of upper bound based on $\alpha$-R\'enyi information]\label{Proposition: decay rate renyi} Assume the hypothesis space is finite and the data samples are i.i.d. Then, the upper bounds based on $\alpha$-R\'enyi information in Theorem~\ref{thm: Minimum upper Bound alpha Renyi} have a convergence rate of $\mathcal{O}(\frac{1}{\sqrt{n}})$. 
\end{proposition}

We can also provide an upper bound based on Sibson’s $\alpha$-mutual information.
\begin{theorem}[Upper bound based on Sibson's $\alpha$ mutual information]\label{Theorem: sibson result}
Assume that the loss function is $\sigma$-sub-Gaussian under distribution $\mu$ for all $w\in \mathcal{W}$ and $\gamma$-sub-Gaussian under $P_{W,Z_i}$, $\forall i=1,\cdots,n$. Then, it holds that:
\begin{align*}
    &|\genb(P_{W|S},\mu)|\leq \frac{1}{n}  \sum_{i=1}^n \sqrt{2(\alpha\sigma^2+(1-\alpha)\gamma^2)\frac{I_{\mathrm{S}}^{\alpha}(W;Z_i)}{\alpha}}.
\end{align*}
\end{theorem}

The upper bound based on $\alpha$-R\'enyi divergence could also be derived using the variational representation of $\alpha$-R\'enyi divergence in \cite{anantharam2018variational}. This approach is applied in \cite{Modak2021Renyi} by considering the sub-Gaussianity under $P_{Z_i}$ and $P_{Z_i|W}$. Our approach is more general, paving the way to offer an upper bound based on $\alpha$-Sibson's mutual information in Theorem~\ref{Theorem: sibson result}, which is derived via ADM.
Since that,
\begin{align}
    I_{\mathrm{S}}^{\alpha}(W;Z_i)&=\min_{Q_W\in\mathcal{P}(\mathcal{W})} \ren(P_{W,Z_i}\| Q_W\otimes \mu)\\
    &\leq \ren(P_{W,Z_i}\| P_W \otimes \mu)
    =I_{\mrR}^\alpha(W;Z_i),
\end{align}
the upper bound in Theorem~\ref{Theorem: sibson result} is tighter than the upper bound in Theorem~\ref{thm: Minimum upper Bound alpha Renyi}. It is worthwhile mentioning that we assume the loss function is $\sigma$-sub-Gaussian under $P_W \otimes \mu$ distribution in Theorem~\ref{thm: Minimum upper Bound alpha Renyi}. However, in Theorem~\ref{Theorem: sibson result}, we consider the loss function is $\sigma$-sub-Gaussian under $\mu$ distribution for all $w\in \mathcal{W}$.

We can also apply generalized Pinsker's inequality~\cite{van2014renyi} to bounded loss functions for bounding the expected generalization error using the $\alpha$-R\'enyi information between data samples, $S$, and hypothesis, $W$.
\begin{proposition}\label{prop: gen_renyi_pinsker's charac}
Consider $\ell(w,z)$ be a bounded loss function i.e. $|\ell(w,z)|\le b$. Then
\begin{align}
    &|\genb(P_{W|S},\mu)|\le 
    \frac{1}{n}\sum_{i=1}^{n}\sqrt{\frac{2b^{2}}{\alpha}I_{\mrR}^\alpha(W;Z_{i})},\quad\forall \alpha\in(0,1].
\end{align}
\end{proposition}
Considering the bounded loss function can help to provide an upper bound based on $\alpha$-Sibson's mutual information between $S$ and $W$ in a similar approach to Proposition~\ref{prop: gen_renyi_pinsker's charac}.

\subsection{Comparison of Proposed Upper Bounds}\label{section: compare}

A summary of upper bounds on expected generalization error under various $\sigma$-sub-Gaussian assumptions is provided in Table~\ref{table: Comparison of Bounds}. 

\begin{table*}[h!]
\caption{Expected Generalization Error Upper Bounds.  We compared our bounds with Mutual information and Lautum information bounds based on the finiteness and the assumption needed for sub-Gaussianity.}
\centering
\resizebox{0.8\textwidth}{!}{\begin{tabular}[h!]{cccc}

 \toprule
 Upper Bound Measure & \makecell{sub-Gaussian \\Assumption} & Bound & Is finite?\\
 \midrule
   Mutual information (\cite{bu2020tightening})
 &
  $P_W\otimes \mu$ 
 &
$\frac{1}{n} \sum_{i=1}^n \sqrt{2\sigma^2I(W;Z_i)}$
&
No\\
\cmidrule(lr{0.5em}){1-4} 
 Lautum information (\cite{gastpar2019Lautum})
 &  
  \makecell{$P_{W,Z_i}$,\\ $\forall i=1,\ldots,n$ } 
 &
$\frac{1}{n} \sum_{i=1}^n \sqrt{2\gamma^2L(W;Z_i)}$
&
No\\
\cmidrule(lr{0.5em}){1-4} 
   \makecell{$\alpha$-$\JDS$ information \\(Proposition~\ref{proposition: Bounded general JS upper})}
 &
 \makecell{$P_{W,Z_i}^{(\alpha)}$,\\ $\forall i=1,\ldots,n$}   
&
 $\frac{1}{n}  \sum_{i=1}^n \sqrt{2\sigma_{(\alpha)}^2\frac{I_{\JDS}^\alpha (W;Z_i)}{\alpha (1-\alpha)}}$
 &
 \makecell{Yes \\($\sigma_{(\alpha)} \sqrt{2\frac{h(\alpha)}{\alpha (1-\alpha)}}$)}\\
\cmidrule(lr{0.5em}){1-4} 
\makecell{$\alpha$-R\'enyi information ($0\leq \alpha <1$) \\(Theorem~\ref{thm: Minimum upper Bound alpha Renyi})}
 &
  \makecell{$P_W\otimes \mu$ and $P_{W,Z_i}$,\\ $\forall i=1,\ldots,n$ }
 &
 $\frac{1}{n}\sum_{i=1}^n \sqrt{2(\alpha\sigma^2+(1-\alpha)\gamma^2)\frac{I_{\mrR}^\alpha(W;Z_i)}{\alpha}}$
 &
 No\\
 \bottomrule
\end{tabular}}

\label{table: Comparison of Bounds}
\end{table*}

\begin{remark}[Bounded loss function]\label{remark: bounded loss}
   The bounded loss function $l:\mathcal{W}\times \mathcal{Z}\rightarrow [a,b]$ is $(\frac{b-a}{2})$-sub-Gaussian under all distributions~\cite{xu2017information}. In fact, for bounded functions, we have,
\begin{align}
  \sigma=\gamma=\sigma_{(\alpha)}=\frac{(b-a)}{2} .
\end{align}
\end{remark}

We next compare the upper bounds based on  $\alpha$-$\JDS$ information, Theorem~\ref{thm: Minimum general js upper Bound}, with the upper bounds based on $\alpha$-R\'enyi information, Theorem~\ref{thm: Minimum upper Bound alpha Renyi}. The next proposition showcases that the  $\alpha$-$\JDS$ information bound can be tighter than the $\alpha$-R\'enyi based upper bound under certain conditions. The proof details are deferred to Appendix~\ref{app: sec compare}.

\begin{proposition}[Comparison of upper bounds based on $\alpha'$-Jensen-Shannon and $\alpha$-R\'enyi information measures] \label{Proposition: alpha renyi to general Jensen-Shannon}
Consider the same assumptions in Theorem~\ref{thm: Minimum general js upper Bound}. Then, it follows that  $\alpha'$-Jensen-Shannon bound given by:
\begin{equation}
    |\genb(P_{W|S},\mu)| \leq \frac{1}{n}  \sum_{i=1}^n \sqrt{2\sigma_{(\alpha')}^2\frac{I_{\JDS}^{\alpha'} (W;Z_i)}{\alpha' (1-\alpha')}}, \quad 0\leq \alpha'\leq 1
\end{equation}
is tighter than the $\alpha$-R\'enyi based upper bound for $ \quad 0\leq \alpha \leq 1$, given by,
\begin{align}
    &|\genb(P_{W|S},\mu)|\leq \frac{1}{n}\sum_{i=1}^n \sqrt{2(\alpha\sigma^2+(1-\alpha)\gamma^2)\frac{I_{\mrR}^\alpha(W;Z_i)}{\alpha}},
\end{align}
provided that $\frac{\alpha h(\alpha')}{(1-\alpha')\alpha'}\leq I_{\mrR}^\alpha(W;Z_i)$ holds for $i=1,\cdots,n$ and $\sigma_{(\alpha')}=\sigma=\gamma$.
\end{proposition}

\begin{remark}
The condition in Proposition~\ref{Proposition: alpha renyi to general Jensen-Shannon}, i.e. $\frac{\alpha h(\alpha')}{(1-\alpha')\alpha'}\leq I_{\mrR}^\alpha(W;Z_i)$, could be tightened by considering $\alpha'=\frac{1}{2}$ and considering the upper bound based on Jensen-Shannon information. 
\end{remark}

\begin{remark}
If we consider $\alpha \rightarrow 1$ and $\alpha'=\frac{1}{2}$ in Proposition~\ref{Proposition: alpha renyi to general Jensen-Shannon}, then the upper bound based on Jensen-Shannon information is tighter than ones based on mutual information~\cite{bu2020tightening} provided that $4\log(2) \leq I(W;Z_i)$ for all $i=1,\cdots,n$ and $\sigma=\sigma_{JS}$.
\end{remark}
\section{Upper Bounds on Excess Risk}\label{sec: excess risk}
This section provides upper bounds on excess risks for regularized empirical risk minimization (ERM) by $\alpha$-R\'enyi divergence or $\alpha$-$\JDS$ divergence.
\subsection{$\alpha$-$\JDS$-Regularized ERM}
It is interesting to consider the regularized ERM with $\alpha$-$\JDS$ information between dataset $S$, and hypothesis $W$,
\begin{equation}\label{eq: js-erm}
        \min_{P_{W|S}}\mbE[L_E(W,S)]+\frac{1}{\beta} I_{\JDS}^{\alpha}(W;S),
\end{equation}
where $\beta>0$ is a parameter that balances fitting and generalization. Since the optimization problem in \eqref{eq: js-erm} is dependent on the data generating distribution, $\mu$, we relax the problem and replace $\alpha$-$\JDS$ information with the $\alpha$-$\JDS$ divergence $\JDS_{\alpha}(P_{W|S}\|Q_{W}|P_S)$, as follows, 
\begin{align}\label{eq: js-erm div}
    \min_{P_{W|S}}\mbE[L_E(W,S)]+\frac{1}{\beta} \JDS_{\alpha}(P_{W|S}\|Q_{W}|P_{S}),
\end{align}
where $Q_W\in\mathcal{P}(\mathcal{W})$ is a prior distribution over parameter space.
\begin{lemma}[Solution existence of $\alpha$-$\JDS$-regularized ERM]
    The optimization problem in \eqref{eq: js-erm div} is a convex optimization problem and has a solution.
    
\end{lemma}
\begin{IEEEproof}
The first term in objective $\mbE[L_E(W,S)]$ is linear in term of $P_{W|S}$ and the second term $\frac{1}{\beta} \JDS_{\alpha}(P_{W|S}\|Q_{W}|P_{S})$ is convex in $P_{W|S}$ for $0<\alpha<1$ due to \cite{sason2018f}. Therefore, a solution exists.
\end{IEEEproof}
Let us define the solution of \eqref{eq: js-erm},
\[P_{W|S}^{\star,\beta,\JDS_{\alpha}}:=\arg \min_{P_{W|S} \in\mathcal{P}(\mathcal{W})}\mbE[L_E(W,S)]+\frac{1}{\beta} \JDS_{\alpha}(P_{W|S}\|Q_{W}|P_{S}).\]
In the following, we provide an upper bound on excess risk under $P_{W|S}^{\star,\beta,\JDS_{\alpha}}$ as a learning algorithm. 
\begin{theorem}[Upper bound on excess risk under $P_{W|S}^{\star,\beta,\JDS_{\alpha}}$] \label{thm: upper true with JSD}
    Assume the bounded loss function, i.e., $|\ell(w,z)|\leq b$ for all $(w,z)\in\mathcal{W}\times \mathcal{Z}$ and $\tilde{L}$-Lipschitz. Then, the following upper bound holds on the excess risk under $P_{W|S}^{\star,\beta,\JDS_{\alpha}}$,
    \[
    \begin{split}
        &\mathcal{E}_r(P_{W|S}^{\star,\beta,\JDS_{\alpha}},\mu)  \leq   \sqrt{\frac{2b^2}{n\alpha(1-\alpha)}\sum_{i=1}^n I_{\JDS}^\alpha(W;Z_i)}\\\quad&+\frac{\tilde{L}\sqrt{d}}{\beta}+\frac{\JDS_{\alpha}(\mathcal{N}(w^\star,\beta^{-1} I_d)\|Q)}{\beta},
    \end{split}
    \]
    where $w^{\star}=\arg\min_{w\in\mathcal{W}} L_\mu(w)$ and $I_d$ is identity matrix with size $d$.
\end{theorem}

\begin{corollary}[Convergence rate of excess risk for under $P_{W|S}^{\star,\beta,\JDS_{\alpha}}$]\label{cor: conv rate of excess risk jsd}
Under the same assumptions in Theorem~\ref{thm: upper true with JSD}, assuming that hypothesis space is finite and $\beta$ is of order $\sqrt{n}$, the following upper bound holds on excess risk of $P_{W|S}^{\star,\beta,\JDS_{\alpha}}$ with convergence rate of $\mathcal{O}(n^{-1/2})$,
\[
    \begin{split}
        &\mathcal{E}_r(P_{W|S}^{\star,\beta,\JDS_{\alpha}},\mu)\leq  \sqrt{\frac{2b^2}{n\alpha(1-\alpha)}\sum_{i=1}^n I_{\JDS}^\alpha(W;Z_i)}\\&\quad+\frac{\tilde{L}\sqrt{d}}{\sqrt{n}}+\frac{h(\alpha)}{\sqrt{n}},
    \end{split}
    \]
\end{corollary}
\begin{remark}[Comparison to the Gibbs algorithm]
    Our convergence rate of the upper bound on the excess risk under $P_{W|S}^{\star,\beta,\JDS_{\alpha}}$ is less than the convergence rate of the upper bound on excess risk under the Gibbs algorithm as the solution of KL-regularized empirical which is  $\mathcal{O}(n^{-1/4})$, \cite[Corollary~3]{xu2017information} and \cite{kuzborskij2019distribution}.  
\end{remark}

\subsection{$\alpha$-R\'enyi-Regularized ERM}
Similarly, it is interesting to consider the regularized ERM with $\alpha$-R\'enyi-information between dataset, $S$, and hypothesis, $W$, for $0<\alpha<1$,
\begin{align}\label{eq: renyi-erm}
    \min_{P_{W|S}}\mbE[L(W,S)]+\frac{1}{\beta} I_{\mrR}^{\alpha}(W;S),
\end{align}
where $\beta>0$ is a parameter that balances fitting and generalization. 

Since the optimization problem in \eqref{eq: renyi-erm} is dependent on the data generating distribution, $\mu$, we propose to relax the problem in \eqref{eq: renyi-erm} by replacing $\alpha$-R\'enyi- information, i.e. $I_{\mrR}^{\alpha}(W;S)$, with $\ren(P_{W|S}\|Q_{W}|P_{S})$ as follows,
\begin{align}\label{eq: RERM_renyi div}
    \min_{P_{W|S}}\mbE[L_E(W,S)]+\frac{1}{\beta} \ren(P_{W|S}\|Q_{W}|P_{S}),
\end{align}
where $Q_W\in\mathcal{P}(\mathcal{W})$.
\begin{lemma}[Solution existence of $\alpha$-R\'enyi-regularized ERM]
The optimization problem considered in \eqref{eq: RERM_renyi div} is a convex optimization problem.
\end{lemma}
\begin{IEEEproof}
The first term in objective $\mbE[L_E(W,S)]$ is linear in term of $P_{W|S}$ and the second term $\frac{1}{\beta} \ren(P_{W|S}\|Q_{W}|P_{S})$ is convex in $P_{W|S}$ for $0<\alpha<1$ due to \cite[Theorem 11]{van2014renyi}. Therefore, a solution exists.
\end{IEEEproof}
Let us define 
\[P_{W|S}^{\star,\beta,\mrR_{\alpha}}:=\arg \min_{P_{W|S} \in\mathcal{P}(\mathcal{W})}\mbE[L_E(W,S)]+\frac{1}{\beta} \ren(P_{W|S}\|Q_{W}|P_{S}),\]
as the solution of convex optimization problem~\eqref{eq: RERM_renyi div}.
\begin{theorem}[Upper bound on excess risk under $P_{W|S}^{\star,\beta,\mrR_{\alpha}}$]\label{thm: upper true with alpha renyi}
    Assume the bounded loss function, i.e., $|\ell(w,z)|\leq b$ for all $(w,z)\in\mathcal{W}\times \mathcal{Z}$ and $\tilde{L}$-Lipschitz. Then, the following upper bound holds on the excess risk under $P_{W|S}^{\star,\beta,\mrR_{\alpha}}$,
    \[
    \begin{split}
       &\mathcal{E}_r(P_{W|S}^{\star,\beta,\mrR_{\alpha}},\mu) \leq   \sqrt{\frac{2b^2}{n\alpha}\sum_{i=1}^n I_{\mrR}^\alpha(W;Z_i)}\\&\quad+\frac{\tilde{L}\sqrt{d}}{\beta}+\frac{\ren(\mathcal{N}(w^\star,\beta^{-1} I_d)\|Q)}{\beta},
    \end{split}
    \]
    where $w^{\star}=\arg\min_{w\in\mathcal{W}} L_\mu(w)$ and $I_d$ is identity matrix with size $d$.
\end{theorem}

\begin{corollary}[Convergence rate of excess risk under $P_{W|S}^{\star,\beta,\mrR_{\alpha}}$]\label{cor: conv rate renyi excess}
    Under the same assumptions in Theorem~\ref{thm: upper true with alpha renyi}, assuming that hypothesis space is finite and $\beta$ is of order $\sqrt{n}$, the following upper bound holds on the excess risk of $P_{W|S}^{\star,\beta,\mrR_{\alpha}}$ with convergence rate of $\mathcal{O}(\log(n)/\sqrt{n})$,
    \[
    \begin{split}
         &\mathcal{E}_r(P_{W|S}^{\star,\beta,\mrR_{\alpha}},\mu)\leq  \sqrt{\frac{2b^2}{n\alpha}\sum_{i=1}^n I_{\mrR}^\alpha(W;Z_i)}\\&\quad+\frac{\tilde{L}\sqrt{d}}{\sqrt{n}}+\frac{1}{2\sqrt{n}}\|w^\star\|_2^2+\frac{d}{4\sqrt{n}}\log\big(n\big)+\frac{d}{2\sqrt{n}(1-\alpha)}\log\big(\alpha\big).
    \end{split}
    \]
\end{corollary}

{\blue
\section{Expected Generalization Error Upper Bounds Under Distribution Mismatch}\label{Sec: mismatch}
In this section, we extend our results in Section~\ref{Sec:Auxiliary distribution Generalization Error} under distribution mismatch, where the training data distribution differs from the test data distribution. All the proof details are deferred to Appendix~\ref{App: Proof of mismatch}.
\begin{proposition}\label{Proposition: alpha general js upper Bound under mismatch}
Assume that the loss function is $\sigma_{(\alpha)}$-sub-Gaussian -- under the distributions $P_{W,Z_i}^{(\alpha)}$ $\forall i=1,\cdots,n$ and $\alpha\mu+(1-\alpha)\mu'$ for all $w\in\mathcal{W}$ -- Then under distribution mismatch \eqref{Eqgen}, it holds $\forall \alpha \in (0,1)$ that:
\begin{align}\label{Eq: Proposition: alpha general js upper Bound under mismatch}
    &|\genb(P_{W|S},\mu,\mu')|\leq \sqrt{2\sigma_{(\alpha)}^2 \frac{\JDS_{\alpha}(\mu' \| \mu)}{\alpha (1-\alpha)}} \\\nn&\quad + \frac{1}{n}  \sum_{i=1}^n \sqrt{2\sigma_{(\alpha)}^2\frac{I_{\JDS}^\alpha (W;Z_i)}{\alpha (1-\alpha)}}, \quad \forall \alpha \in (0,1).
\end{align}
\end{proposition}

\begin{proposition}\label{Prop: upper Bound alpha Renyi under mismatch} Assume that the loss function is $\sigma$-sub-Gaussian under distributions $\mu$ and $\mu'$ for all $w\in\mathcal{W}$ and also $\gamma$-sub-Gaussian under $P_{W,Z_i}$ $\forall i=1,\cdots,n$. The following upper bound for $\forall \alpha \in (0,1)$ holds,

\begin{align}\label{Eq: Prop: upper Bound alpha Renyi under mismatch}
    &|\genb(P_{W|S},\mu,\mu')|\leq \sqrt{2(\alpha\sigma^2+(1-\alpha)\gamma^2)\frac{\ren(\mu'\|\mu)}{\alpha}}\\\nn&\quad +\frac{1}{n}  \sum_{i=1}^n \sqrt{2(\alpha\sigma^2+(1-\alpha)\gamma^2)\frac{I_{\mrR}^\alpha(W;Z_i)}{\alpha}}.
\end{align}
\end{proposition}

The mismatch between the test and training samples distributions is characterised in \cite[Theorem~5]{masiha2021learning} as KL divergence between test and training samples distributions, i.e., $\KLr(\mu'\|\mu)$. However, assuming that the loss function is $\sigma_{(\alpha)}$-sub-Gaussian under $\alpha\mu+(1-\alpha)\mu'$ for all $w\in\mathcal{W}$, Proposition~\ref{Proposition: alpha general js upper Bound under mismatch} allows us to explain the distributional mismatch in terms of  $\alpha$-Jensen-Shannon divergence, which is finite.

In Proposition~\ref{Prop: upper Bound alpha Renyi under mismatch}, the distributional mismatch is presented in terms of $\alpha$-R\'enyi divergence, i.e., $\ren(\mu'\|\mu)$. If $\mu'$ is not absolutely continuous with respect to $\mu$, then we have $\KLr(\mu'\|\mu)=\infty$. However, for $\alpha$-R\'enyi divergence ($0<\alpha<1$), it suffices that the mutual singularity~\cite{van2014renyi}, i.e., $\mu'\indep\mu$, does not hold, which is a less restrictive condition about $\mu'$ compared to the absolutely continuity condition.

Similar to Remark~\ref{remark: bounded loss}, the sub-Gaussianity assumptions in Propositions~\ref{Proposition: alpha general js upper Bound under mismatch} and \ref{Prop: upper Bound alpha Renyi under mismatch} hold for bounded loss functions.

}

\section{Numerical Example}
In this section, we illustrate that some of our proposed bounds can be tighter than existing ones in a simple toy example. We consider the  $\alpha$-$\JDS$ and $\alpha$-R\'enyi information only. Our example setting involves the estimation of the mean of a Gaussian random variable $Z \sim \mathcal{N}(\beta,\sigma^2)$ based on two i.i.d. samples $Z_1$ and $Z_2$. We consider the hypothesis (estimate) given by $W=tZ_1+(1-t)Z_2$ for $0<t<1$. We also consider the loss function given by $\ell(w,z)=\min((w-z)^2,c^2)$.

{\blue Due to the fact that the loss function is bounded within the interval $[0,c^2]$, then it is $\frac{c^2}{2}$-sub-Gaussian under all distributions.} Therefore, we can apply the expected generalization error upper bounds based on mutual information,  $\alpha$-$\JDS$ information and $\alpha$-R\'enyi information $\forall \alpha \in (0,1)$ as follows:
\begin{align}\label{Eq: upper MI bound Sim}
    &\genb(P_{W|Z_1,Z_2},P_Z)\leq \frac{c^2}{4}\big(\sqrt{2I(W;Z_1)}+\sqrt{2I(W;Z_2)}\big),\\\label{Eq: upper JS bound Sim}
    &\genb(P_{W|Z_1,Z_2},P_Z)\leq \frac{c^2}{4}\Big(\sqrt{2\frac{I_{\JDS}^\alpha(W;Z_1)}{\alpha(1-\alpha)}}+\sqrt{2\frac{I_{\JDS}^\alpha(W;Z_2)}{\alpha(1-\alpha)}}\Big),\\\label{Eq: upper renyi bound Sim}
    &\genb(P_{W|Z_1,Z_2},P_Z)\leq \frac{c^2}{4}\Big(\sqrt{2\frac{I_{\textit{R}}^\alpha(W;Z_1)}{\alpha}}+\sqrt{2\frac{I_{\textit{R}}^\alpha(W;Z_2)}{\alpha}}\Big).
\end{align}
It can be immediately shown that $W \sim \mathcal{N}(\beta,\sigma^2(t^2+(1-t)^2))$ and $(W,Z_1)$ and $(W,Z_2)$ are jointly Gaussian with correlation coefficients $\rho_1=\frac{t}{\sqrt{t^2+(1-t)^2}}$ and $\rho_2=\frac{(1-t)}{\sqrt{t^2+(1-t)^2}}$. Therefore, it can be shown that the mutual information appearing above is given by $I(W;Z_1)=-\frac{1}{2}\log(1-\rho_1^2)$ and $I(W;Z_2)=-\frac{1}{2}\log(1-\rho_2^2)$. In contrast, the  $\alpha$-$\JDS$ information appearing above can be computed via an extension of entropic-based formulation of the Jensen-Shannon measure as follows \cite{lin1991divergence}:
\begin{align}
    &I_{\JDS}(W;Z_i)=\\\nn&\quad h\left(P_{W,Z_i}^{(\alpha)}\right)-(\alpha h(P_W)+\alpha h(P_{Z_i})+(1-\alpha)h(P_{Z_i,W})),
\end{align}
-- with $h (\cdot)$ denoting the differential entropy -- where
\begin{align*}
 &h(P_{Z_i})=\frac{1}{2}\log(2\pi\sigma^2),\\
  &h(P_W)=\frac{1}{2}\log(2\pi\sigma^2(t^2+(1-t)^2)),\\
  &h(P_{W,Z_i})=\log(2\pi\sigma^2(t^2+(1-t)^2)(1-\rho_i^2)),
\end{align*}
 whereas $h\left(P_{W,Z_i}^{(\alpha)}\right)$ can be computed numerically.

\begin{figure}[!htbp]
    \centering
    \includegraphics[scale=0.22]{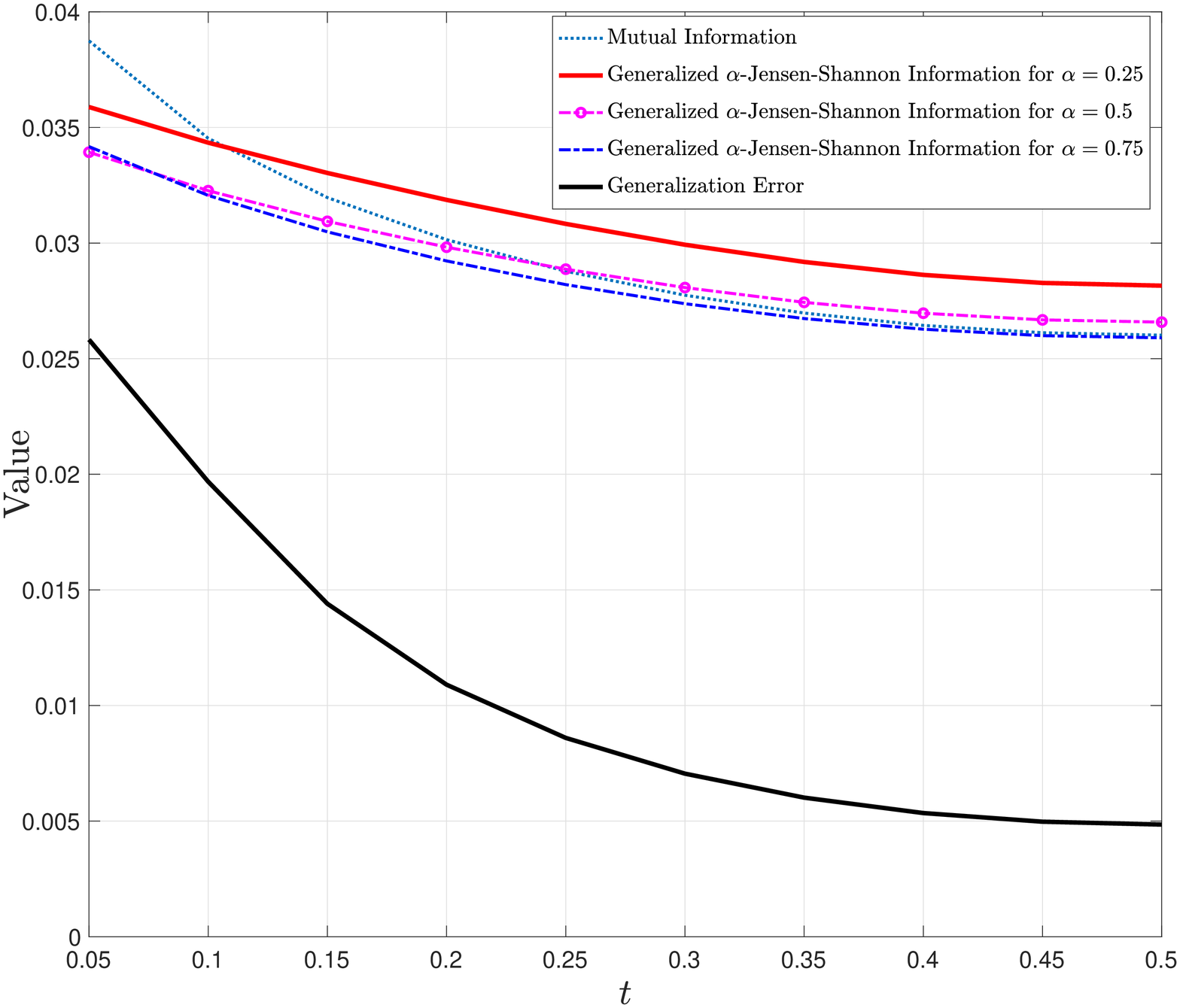}
    \caption{True generalization error,  $\alpha$-$\JDS$ based bound for $\alpha=0.25,0.5,0.75$, and Mutual Information based bound.}
    \label{fig:Upper bounds compare GE with Gjensen}
\end{figure}

Fig.\ref{fig:Upper bounds compare GE with Gjensen} depicts the true generalization error, the mutual information based bound in \eqref{Eq: upper MI bound Sim}, and the  $\alpha$-$\JDS$ information based bound for $\alpha=0.25,0.5,0.75$ in \eqref{Eq: upper JS bound Sim} for values of $t\in(0,0.5]$, considering $\sigma^2=1, 10$, $\mu=1$, $c=\frac{\sigma}{4}$. 

\begin{figure}[!htbp]
    \centering
    \includegraphics[scale=0.23]{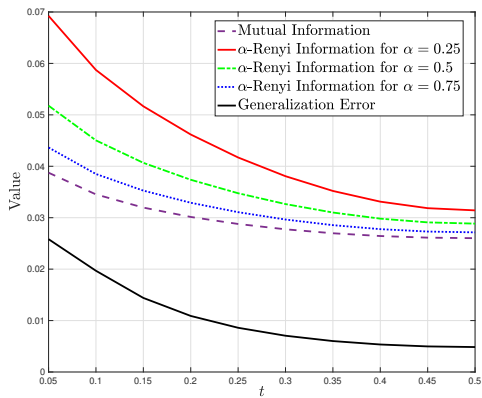}
    \caption{True generalization error, $\alpha$-R\'enyi based bound for $\alpha=0.25,0.5,0.75$, and Mutual Information based bound.}
    \label{fig:Upper bounds compare GE with Renyi}
\end{figure}

It can be seen that for $\alpha=0.75$ the  $\alpha$-$\JDS$ information bound is tighter than the mutual information bound. For $\alpha=0.5$, which is equal to traditional Jensen-Shannon information, if we consider $t<0.25$ then the Jensen-Shannon information bound is tighter than the mutual information bound; in contrast, for $t>0.25$, the mutual information bound is slightly better than the Jensen-Shannon information bound. This showcases that our proposed bounds can be tighter than existing ones in some regimes.
Fig.\ref{fig:Upper bounds compare GE with Renyi} also depicts the true generalization error, the mutual information based bound in \eqref{Eq: upper MI bound Sim}, and the $\alpha$-R\'enyi information based bound for $\alpha=0.25,0.5,0.75$ in \eqref{Eq: upper renyi bound Sim}. It can be seen that the $\alpha$-R\'enyi based bound is looser than the mutual information based bound. {\blue In our experiment setup, when $t\to 0$ (or $t\to 1$), we have $I(W;Z_2)\to\infty$ (or $I(W;Z_1)\to\infty$). However, the $\alpha$-R\'enyi based bound is finite.}
\section{Conclusion and Future Works}\label{Conc}
We have presented the Auxiliary Distribution Method, a novel approach for deriving information-theoretic upper bounds on the generalization error within the context of supervised learning problems. Our method offers the flexibility to recover existing bounds while also enabling the derivation of new bounds grounded in the  $\alpha$-$\JDS$ and $\alpha$-R\'enyi information measures. Notably, our upper bounds, which are rooted in the  $\alpha$-$\JDS$ information measure, are finite, in contrast to mutual information-based bounds. Moreover, our upper bound based on $\alpha$-R\'enyi information, for $\alpha \in (0,1)$, remains finite when considering a deterministic learning process. An intriguing observation is that our newly introduced  $\alpha$-$\JDS$ information measure can, in certain regimes, yield tighter bounds compared to existing approaches. We also discuss the existence of algorithms under $\alpha$-$\JDS$-regularized and $\alpha$-R\'enyi-regularized empirical risk minimization problems and provide upper bounds on excess risk of these algorithms, where the upper bound on the excess risk under $\alpha$-$\JDS$-regularized empirical risk minimization is tighter than other well-known upper bounds on excess risk. {\blue Furthermore, we provide an upper bound on generalization error in a mismatch scenario, where the distributions of test and training datasets are different, via our auxiliary distribution method.}

As a direction for future research, we propose extending our bounds to the PAC-Bayesian framework, leveraging the $\alpha$-$\JDS$ and $\alpha$-R\'enyi divergences for $0 < \alpha < 1$. Additionally, the conditional technique based on individual sample measures, as described in~\cite{zhou2020individually}, could be applied to improve the effectiveness of our upper bounds.

\section*{Acknowledgment}
Gholamali Aminian is supported in part by the Royal Society Newton International Fellowship, grant no. NIF\textbackslash R1 \textbackslash192656, the UKRI Prosperity Partnership Scheme (FAIR) under the EPSRC Grant EP/V056883/1, and the Alan Turing Institute. Saeed Masiha worked on this project at Sharif university of Technology before joining EPFL University.

\bibliographystyle{ieeetr}
\bibliography{Refs}

\begin{thebibliography}{10}

\bibitem{aminian2021Jensen}
G.~Aminian, L.~Toni, and M.~R.~D. Rodrigues, ``Jensen-shannon information based
  characterization of the generalization error of learning algorithms,'' in
  {\em 2020 IEEE Information Theory Workshop (ITW)}, pp.~1--5, 2021.

\bibitem{vapnik1999overview}
V.~N. Vapnik, ``An overview of statistical learning theory,'' {\em IEEE
  transactions on neural networks}, vol.~10, no.~5, pp.~988--999, 1999.

\bibitem{bousquet2002stability}
O.~Bousquet and A.~Elisseeff, ``Stability and generalization,'' {\em Journal of
  machine learning research}, vol.~2, no.~Mar, pp.~499--526, 2002.

\bibitem{xu2012robustness}
H.~Xu and S.~Mannor, ``Robustness and generalization,'' {\em Machine learning},
  vol.~86, no.~3, pp.~391--423, 2012.

\bibitem{mcallester2003pac}
D.~A. McAllester, ``Pac-bayesian stochastic model selection,'' {\em Machine
  Learning}, vol.~51, no.~1, pp.~5--21, 2003.

\bibitem{russo2019much}
D.~Russo and J.~Zou, ``How much does your data exploration overfit? controlling
  bias via information usage,'' {\em IEEE Transactions on Information Theory},
  vol.~66, no.~1, pp.~302--323, 2019.

\bibitem{xu2017information}
A.~Xu and M.~Raginsky, ``Information-theoretic analysis of generalization
  capability of learning algorithms,'' in {\em Advances in Neural Information
  Processing Systems}, pp.~2524--2533, 2017.

\bibitem{bu2020tightening}
Y.~Bu, S.~Zou, and V.~V. Veeravalli, ``Tightening mutual information-based
  bounds on generalization error,'' {\em IEEE Journal on Selected Areas in
  Information Theory}, vol.~1, no.~1, pp.~121--130, 2020.

\bibitem{asadi2018chaining}
A.~Asadi, E.~Abbe, and S.~Verd{\'u}, ``Chaining mutual information and
  tightening generalization bounds,'' in {\em Advances in Neural Information
  Processing Systems}, pp.~7234--7243, 2018.

\bibitem{asadi2020chaining}
A.~R. Asadi and E.~Abbe, ``Chaining meets chain rule: Multilevel entropic
  regularization and training of neural networks,'' {\em Journal of Machine
  Learning Research}, vol.~21, no.~139, pp.~1--32, 2020.

\bibitem{esposito2019generalization}
A.~R. Esposito, M.~Gastpar, and I.~Issa, ``Generalization error bounds via
  r{\'e}nyi-, f-divergences and maximal leakage,'' {\em IEEE Transactions on
  Information Theory}, vol.~67, no.~8, pp.~4986--5004, 2021.

\bibitem{Modak2021Renyi}
E.~Modak, H.~Asnani, and V.~M. Prabhakaran, ``Rényi divergence based bounds on
  generalization error,'' in {\em 2021 IEEE Information Theory Workshop (ITW)},
  pp.~1--6, 2021.

\bibitem{lopez2018generalization}
A.~T. Lopez and V.~Jog, ``Generalization error bounds using wasserstein
  distances,'' in {\em 2018 IEEE Information Theory Workshop (ITW)}, pp.~1--5,
  IEEE, 2018.

\bibitem{wang2019information}
H.~Wang, M.~Diaz, J.~C.~S. Santos~Filho, and F.~P. Calmon, ``An
  information-theoretic view of generalization via wasserstein distance,'' in
  {\em 2019 IEEE International Symposium on Information Theory (ISIT)},
  pp.~577--581, IEEE, 2019.

\bibitem{galvez2021tighter}
B.~R. G{\'a}lvez, G.~Bassi, R.~Thobaben, and M.~Skoglund, ``Tighter expected
  generalization error bounds via wasserstein distance,'' in {\em Advances in
  Neural Information Processing Systems}, 2021.

\bibitem{aminian2022tighter}
G.~Aminian, Y.~Bu, G.~Wornell, and M.~Rodrigues, ``Tighter expected
  generalization error bounds via convexity of information measures,'' in {\em
  IEEE International Symposium on Information Theory (ISIT)}, 2022.

\bibitem{steinke2020reasoning}
T.~Steinke and L.~Zakynthinou, ``Reasoning about generalization via conditional
  mutual information,'' in {\em Conference on Learning Theory}, pp.~3437--3452,
  PMLR, 2020.

\bibitem{zhou2020individually}
R.~Zhou, C.~Tian, and T.~Liu, ``Individually conditional individual mutual
  information bound on generalization error,'' {\em IEEE Transactions on
  Information Theory}, pp.~1--1, 2022.

\bibitem{hafez2020conditioning}
H.~Hafez-Kolahi, Z.~Golgooni, S.~Kasaei, and M.~Soleymani, ``Conditioning and
  processing: Techniques to improve information-theoretic generalization
  bounds,'' {\em Advances in Neural Information Processing Systems}, vol.~33,
  2020.

\bibitem{aminian2021exact}
G.~Aminian$^*$, Y.~Bu$^*$, L.~Toni, M.~Rodrigues, and G.~Wornell, ``An exact
  characterization of the generalization error for the {G}ibbs algorithm,''
  {\em Advances in Neural Information Processing Systems}, vol.~34, 2021.

\bibitem{masiha2021learning}
M.~S. Masiha, A.~Gohari, M.~H. Yassaee, and M.~R. Aref, ``Learning under
  distribution mismatch and model misspecification,'' in {\em IEEE
  International Symposium on Information Theory (ISIT)}, 2021.

\bibitem{mansour2009domain}
Y.~Mansour, M.~Mohri, and A.~Rostamizadeh, ``Domain adaptation: Learning bounds
  and algorithms,'' {\em arXiv preprint arXiv:0902.3430}, 2009.

\bibitem{wang2018theoretical}
Z.~Wang, ``Theoretical guarantees of transfer learning,'' 2018.

\bibitem{wu2020information}
X.~Wu, J.~H. Manton, U.~Aickelin, and J.~Zhu, ``Information-theoretic analysis
  for transfer learning,'' in {\em 2020 IEEE International Symposium on
  Information Theory (ISIT)}, pp.~2819--2824, IEEE, 2020.

\bibitem{englesson2021generalized}
E.~Englesson and H.~Azizpour, ``Generalized jensen-shannon divergence loss for
  learning with noisy labels,'' {\em Advances in Neural Information Processing
  Systems}, vol.~34, pp.~30284--30297, 2021.

\bibitem{goodfellow2014generative}
I.~Goodfellow, J.~Pouget-Abadie, M.~Mirza, B.~Xu, D.~Warde-Farley, S.~Ozair,
  A.~Courville, and Y.~Bengio, ``Generative adversarial nets,'' in {\em
  Advances in neural information processing systems}, pp.~2672--2680, 2014.

\bibitem{melville2005active}
P.~Melville, S.~M. Yang, M.~Saar-Tsechansky, and R.~Mooney, ``Active learning
  for probability estimation using jensen-shannon divergence,'' in {\em
  European conference on machine learning}, pp.~268--279, Springer, 2005.

\bibitem{choi2003feature}
E.~Choi and C.~Lee, ``Feature extraction based on the bhattacharyya distance,''
  {\em Pattern Recognition}, vol.~36, no.~8, pp.~1703--1709, 2003.

\bibitem{coleman1979image}
G.~B. Coleman and H.~C. Andrews, ``Image segmentation by clustering,'' {\em
  Proceedings of the IEEE}, vol.~67, no.~5, pp.~773--785, 1979.

\bibitem{topsoe2007information}
F.~Tops{\o}e, ``Information theory at the service of science,'' in {\em
  Entropy, Search, Complexity}, pp.~179--207, Springer, 2007.

\bibitem{topsoe2000some}
F.~Topsoe, ``Some inequalities for information divergence and related measures
  of discrimination,'' {\em IEEE Transactions on Information Theory}, vol.~46,
  no.~4, pp.~1602--1609, 2000.

\bibitem{cover1999elements}
T.~M. Cover, {\em Elements of information theory}.
\newblock John Wiley \& Sons, 1999.

\bibitem{nielsen2020generalization}
F.~Nielsen, ``On a generalization of the jensen-shannon divergence and the
  jensen-shannon centroid,'' {\em Entropy}, vol.~22, no.~2, p.~221, 2020.

\bibitem{lin1991divergence}
J.~Lin, ``Divergence measures based on the shannon entropy,'' {\em IEEE
  Transactions on Information Theory}, vol.~37, no.~1, pp.~145--151, 1991.

\bibitem{van2014renyi}
T.~Van~Erven and P.~Harremos, ``R{\'e}nyi divergence and kullback-leibler
  divergence,'' {\em IEEE Transactions on Information Theory}, vol.~60, no.~7,
  pp.~3797--3820, 2014.

\bibitem{kailath1967divergence}
T.~Kailath, ``The divergence and bhattacharyya distance measures in signal
  selection,'' {\em IEEE transactions on communication technology}, vol.~15,
  no.~1, pp.~52--60, 1967.

\bibitem{palomar2008lautum}
D.~P. Palomar and S.~Verd{\'u}, ``Lautum information,'' {\em IEEE Transactions
  on Information Theory}, vol.~54, no.~3, pp.~964--975, 2008.

\bibitem{sason2016f}
I.~Sason and S.~Verd{\'u}, ``$ f $-divergence inequalities,'' {\em IEEE
  Transactions on Information Theory}, vol.~62, no.~11, pp.~5973--6006, 2016.

\bibitem{verdu2015alpha}
S.~Verd{\'u}, ``$\alpha$-mutual information,'' in {\em 2015 Information Theory
  and Applications Workshop (ITA)}, pp.~1--6, IEEE, 2015.

\bibitem{boucheron2013concentration}
S.~Boucheron, G.~Lugosi, and P.~Massart, {\em Concentration inequalities: A
  nonasymptotic theory of independence}.
\newblock Oxford university press, 2013.

\bibitem{dupuis2011weak}
P.~Dupuis and R.~S. Ellis, {\em A weak convergence approach to the theory of
  large deviations}, vol.~902.
\newblock John Wiley \& Sons, 2011.

\bibitem{gastpar2019Lautum}
M.~Gastpar, A.~R. Esposito, and I.~Issa, ``Information measures, learning and
  generalization,'' {\em 5th London Symposium on Information Theory}, 2019.

\bibitem{topsoe2002inequalities}
F.~Topsoe, ``Inequalities for the jensen-shannon divergence,'' {\em Draft
  available at http://www. math. ku. dk/topsoe}, 2002.

\bibitem{raginsky2016information}
M.~Raginsky, A.~Rakhlin, M.~Tsao, Y.~Wu, and A.~Xu, ``Information-theoretic
  analysis of stability and bias of learning algorithms,'' in {\em 2016 IEEE
  Information Theory Workshop (ITW)}, pp.~26--30, IEEE, 2016.

\bibitem{anantharam2018variational}
V.~Anantharam, ``A variational characterization of r{\'e}nyi divergences,''
  {\em IEEE Transactions on Information Theory}, vol.~64, no.~11,
  pp.~6979--6989, 2018.

\bibitem{sason2018f}
I.~Sason, ``On f-divergences: Integral representations, local behavior, and
  inequalities,'' {\em Entropy}, vol.~20, no.~5, p.~383, 2018.

\bibitem{kuzborskij2019distribution}
I.~Kuzborskij, N.~Cesa-Bianchi, and C.~Szepesv{\'a}ri, ``Distribution-dependent
  analysis of gibbs-erm principle,'' in {\em Conference on Learning Theory},
  pp.~2028--2054, PMLR, 2019.

\bibitem{topsoe2003jenson}
F.~Tops{\o}e, ``Jenson-shannon divergence and norm-based measures of
  discrimination and variation,'' {\em preprint}, 2003.

\bibitem{gil2013renyi}
M.~Gil, F.~Alajaji, and T.~Linder, ``R{\'e}nyi divergence measures for commonly
  used univariate continuous distributions,'' {\em Information Sciences},
  vol.~249, pp.~124--131, 2013.

\end{thebibliography}
\clearpage
\newpage
\appendices

\section{Proof of Section~\ref{subsection: Jensen-Shannon}}\label{app: proof of jsd}

\begin{IEEEproof}[Proof of Lemma~\ref{lemma: Jensen-Shannon equality}]
\begin{align}
    &\alpha \KLr(P_W\otimes P_{Z_i}\|\widehat{P}_{W,Z_i})+(1-\alpha)\KLr(P_{W,Z_i}\|\widehat{P}_{W,Z_i})\\
    &=\int_{\mathcal{W} \times \mathcal{Z}}\alpha (\mrd P_W\otimes \mrd P_{Z_i})\log(\mrd P_W\otimes \mrd P_{Z_i})\\\nn&\quad+\int_{\mathcal{W} \times \mathcal{Z}}(1-\alpha)\mrd P_{W,Z_i}\log(\mrd P_{W,Z_i})\\\nn
    &\quad-\int_{\mathcal{W} \times \mathcal{Z}}((\alpha (\mrd P_W\otimes \mrd P_{Z_i})+(1-\alpha)\mrd P_{W,Z_i})\log(d\widehat{P}_{W,Z_i}))\\
    &=\int_{\mathcal{W} \times \mathcal{Z}}\alpha (\mrd P_W\otimes \mrd P_{Z_i})\log(\mrd P_W\otimes \mrd P_{Z_i})\\\nn&\quad+\int_{\mathcal{W} \times \mathcal{Z}}(1-\alpha)\mrd P_{W,Z_i}\log(\mrd P_{W,Z_i})
    -\mrd P_{W,Z_i}^{(\alpha)}\log(d\widehat{P}_{W,Z_i}) \\\nn&\quad+\int_{\mathcal{W} \times \mathcal{Z}} \mrd P_{W,Z_i}^{(\alpha)}\log(\mrd P_{W,Z_i}^{(\alpha)})  - \mrd P_{W,Z_i}^{(\alpha)}\log(\mrd P_{W,Z_i}^{(\alpha)})\\
    &=I_{\JDS}^\alpha (W;Z_i)+\KLr(P_{W,Z_i}^{(\alpha)}\|\widehat{P}_{W,Z_i}).
\end{align}

\end{IEEEproof}

\begin{IEEEproof}[Proof of Theorem~\ref{thm: Minimum general js upper Bound}]
As shown in \cite{topsoe2003jenson}, and by considering the Lemma~\ref{lemma: Jensen-Shannon equality} we have
\begin{align}
&\min_{\widehat{P}_{W,Z_i}}\alpha \KLr(P_W\otimes\mu\|\widehat{P}_{W,Z_i})+(1-\alpha)\KLr(P_{W,Z_i}\|\widehat{P}_{W,Z_i})=\\\nn
&\min_{\widehat{P}_{W,Z_i}} I_{\JDS}^\alpha (W;Z_i)+\KLr(P_{W,Z_i}^{(\alpha)}\|\widehat{P}_{W,Z_i}).
\end{align}
As we have $0\leq \KLr(P_{W,Z_i}^{(\alpha)}\|\widehat{P}_{W,Z_i})$, therefore, the minimum of \eqref{Eq: Proposition: General Upper Bounds} is achieved with $\widehat{P}_{W,Z_i}=P_{W,Z_i}^{(\alpha)}$. Now, considering $\widehat{P}_{W,Z_i}=P_{W,Z_i}^{(\alpha)}$ in Proposition~\ref{Prop: Avergae Direct Upper bound}, completes the proof.
\end{IEEEproof}


\begin{IEEEproof}[Proof of Proposition~\ref{Proposition: decay rate Jensen-Shannon}]
    
Using \eqref{Eq: genaral jensen inequality},
\begin{align}
    I_{\JDS}^\alpha(W;Z_i)\leq (1-\alpha)I(W;Z_i),
\end{align}
we have:
\begin{align}
    |\genb(P_{W|S},\mu)|&\leq \frac{1}{n}  \sum_{i=1}^n \sqrt{2\sigma_{(\alpha)}^2\frac{I_{\JDS}^\alpha (W;Z_i)}{\alpha (1-\alpha)}}\\
    &\leq\frac{1}{n}  \sum_{i=1}^n \sqrt{2\sigma_{(\alpha)}^2\frac{I (W;Z_i)}{\alpha }}\\
    &\leq \sqrt{2\sigma_{(\alpha)}^2\frac{\sum_{i=1}^n I(W;Z_i)}{\alpha n}}\\
    &\leq \sqrt{2\sigma_{(\alpha)}^2\frac{I(W;S)}{\alpha n}}\\
    &\leq \sqrt{2\sigma_{(\alpha)}^2\frac{H(W)}{\alpha n}},
\end{align}
where the final result would follow from the finite hypothesis space.
\end{IEEEproof}

\begin{IEEEproof}[Proof of Proposition~\ref{proposition: Bounded general JS upper}]
This proposition follows from  the fact that $I_\JDS^\alpha(W,Z_i)\leq h(\alpha)$ for $i=1,\cdots,n$.

We prove that $I_\JDS^\alpha(W,Z_i)\leq h(\alpha)$.
\begin{align}
  &I_\JDS^\alpha(W,Z_i)=\\\nn&\quad
  \alpha \KLr(P_W\otimes P_{Z_i}\|P_{W,Z_i}^{(\alpha)})+(1-\alpha)\KLr(P_{W,Z_i}\| P_{W,Z_i}^{(\alpha)})\\
  &=\alpha \int_{\mathcal{W}\times \mathcal{Z}}\mrd P_W\otimes \mrd P_{Z_i} \log\left(\frac{\mrd P_W\otimes \mrd P_{Z_i}}{\mrd P_{W,Z_i}^{(\alpha)}}\right)\\\nn&\quad+(1-\alpha)\int_{\mathcal{W}\times \mathcal{Z}}\mrd P_{W,Z_i} \log\left(\frac{\mrd P_{W,Z_i}}{\mrd P_{W,Z_i}^{(\alpha)}}\right)\\
  &\leq\alpha \int_{\mathcal{W}\times \mathcal{Z}}\mrd P_W\otimes \mrd P_{Z_i} \log\left(\frac{\mrd P_W\otimes \mrd P_{Z_i}}{\alpha (\mrd P_W\otimes \mrd P_{Z_i})}\right) \\\nn&\quad+(1-\alpha)\int_{\mathcal{W}\times \mathcal{Z}}\mrd P_{W,Z_i} \log\left(\frac{\mrd P_{W,Z_i}}{(1-\alpha)\mrd P_{W,Z_i}}\right)\\
  &=-\alpha\log(\alpha)-(1-\alpha)\log(1-\alpha)\\
  &=h(\alpha).
\end{align}

\end{IEEEproof}
\begin{IEEEproof}[Proof of Corollary~\ref{corollary: const upper bound optimum}]
We first compute the derivative of $ \frac{h(\alpha)}{\alpha (1-\alpha)}$ with respect to $\alpha \in (0,1)$
\begin{align}
    \frac{d \frac{h(\alpha)}{\alpha (1-\alpha)}}{d \alpha}=\frac{\log(1-\alpha)}{\alpha^2}-\frac{\log(\alpha)}{(1-\alpha)^2}.
\end{align}
Now for $\alpha=\frac{1}{2}$, we have $\frac{d \frac{h(\alpha)}{\alpha (1-\alpha)}}{d \alpha}=0$.
\end{IEEEproof}

\section{Proofs of Section~\ref{subsection: Renyi bounds}}\label{app: proof of renyi}
\begin{IEEEproof}[Proof of Proposition~\ref{Prop: reverse upper bound sigma subgaussian}]
   Consider arbitrary auxiliary distributions $\{\widehat{P}_{W,Z_i}\}_{i=1}^n$ defined on $\mathcal{W}\times \mathcal{Z}$. Then,
\begin{align}\label{Eq: Ex 3}
    &\genb(P_{W|S},\mu)=\mbE_{P_{W}P_{S}}[L_E(W,S)]-\mbE_{P_{W,S}}[L_E(W,S)]\nn\\
    &=\frac{1}{n}\sum_{i=1}^n \mbE_{P_{W}P_{Z_i}}[\ell(W,Z_i)]- \mbE_{P_{W Z_i}}[\ell(W,Z_i)] \\
    &\leq \frac{1}{n}\sum_{i=1}^n \left|\mbE_{P_{W}P_{Z_i}}[\ell(W,Z_i)]- \mbE_{P_{W Z_i}}[\ell(W,Z_i)]\right|\\
    &\leq \frac{1}{n}\sum_{i=1}^n \bigl|\mbE_{\widehat{P}_{W,Z_i}}[\ell(W,Z_i)]- \mbE_{P_{W Z_i}}[\ell(W,Z_i)]\bigr|\\\nonumber&\quad\quad
    +\bigl|\mbE_{\widehat{P}_{W,Z_i}}[\ell(W,Z_i)]-\mbE_{P_{W}P_{Z_i}}[\ell(W,Z_i)]\bigr|.
\end{align}

Using the assumption that loss function $\ell(w,z_i)$ is $\gamma^2$-sub-Gaussian under distribution $P_{W,Z_i}$ and Donsker-Varadhan representation we have:
\begin{align}\label{10013}
&\lambda\left(\mbE_{\widehat{P}_{W,Z_i}}[\ell(W,Z_i)]- \mbE_{P_{W Z_i}}[\ell(W,Z_i)]\right ) \le\\\nn&\quad \KLr(\widehat{P}_{W,Z_i}\|P_{W Z_i})+\frac{\lambda^2\gamma^{2}}{2},\quad\forall\lambda\in \mathbb{R}.
\end{align}

Using the assumption that $\ell(w,Z)$ is $\sigma^{2}$-sub-Gaussian under $P_W \otimes P_{Z_i}$, and again Donsker-Varadhan representation we have:
\begin{align}\label{1002}
 &\lambda'\left( \mbE_{\widehat{P}_{W,Z_i}}[\ell(W,Z_i)]-\mbE_{P_{W}P_{Z_i}}[\ell(W,Z_i)]\right)\le\\\nn&\quad \KLr(\widehat{P}_{W,Z_i}\|P_{W}P_{Z_i})+\frac{{\lambda'}^2\sigma^{2}}{2}.\quad\forall\lambda'\in \mathbb{R}
\end{align}
Note that $\mbE_{P_{W}P_{Z_i}}[\ell(W,Z_i)-\mbE_{P_{Z_i}}[\ell(W,Z_i)]]=0$.

Now if we consider $\lambda>0$, then we choose $\lambda'=\frac{\alpha}{\alpha-1} \lambda$. Hence we have

\begin{align}\label{1003}
&\mbE_{\widehat{P}_{W,Z_i}}[\ell(W,Z_i)-\mbE_{P_{W,Z_{i}}}[\ell(W,Z_{i})]]\le \\\nn&\quad \frac{\KLr(\widehat{P}_{W,Z_i}\|P_{W Z_i})}{\lambda}+\frac{\lambda\gamma^{2}}{2},\quad\forall\lambda\in \mathbb{R^+}.
\end{align}
Using the assumption that $\ell(w,Z)$ is $\sigma^{2}$-sub-Gaussian and again Donsker-Varadhan representation,
\begin{align}\label{1004}
&- \mbE_{\widehat{P}_{W,Z_i}}[\ell(W,Z_i)-\mbE_{P_{W}P_{Z_i}}[\ell(W,Z_i)]]\le \\\nn&\quad\frac{\KLr(\widehat{P}_{W,Z_i}\|P_{W}P_{Z_i})}{|\lambda'|}+\frac{|\lambda'|\sigma^{2}}{2},\quad\forall\lambda'\in \mathbb{R^-}.
\end{align}

Now sum up two Inequalities \eqref{1003} and \eqref{1004}, to obtain
\begin{align}\label{Eq: Ex 51}
    &\mbE_{P_W P_{Z_i}}[\ell(W,Z_i)]-\mbE_{P_{W Z_i}}[\ell(W,Z_i)]\le\\\nn
    &\frac{\alpha \KLr(\widehat{P}_{W,Z_i}\|P_{W,Z_i})+(1-\alpha)\KLr(\widehat{P}_{W,Z_i}\|Q_{W}P_{Z_i})}{\alpha\lambda}+\\\nn&\quad\frac{\lambda\gamma^{2}}{2}+\frac{\lambda\frac{\alpha}{1-\alpha}\sigma^{2}}{2},\quad\forall\lambda\in\mathbb{R^+}.
\end{align}
Considering \eqref{Eq: Ex 51}, we have a nonnegative parabola in $\lambda$, whose discriminant must be nonpositive, and we have:
\begin{align}
 & \left | \mbE_{P_W P_{Z_i}}[\ell(W,Z_i)]-\mbE_{P_{W Z_i}}[\ell(W,Z_i)]\right|\le\\\nn&\quad \sqrt{2(\alpha\sigma^2+(1-\alpha)\gamma^2)\frac{\left(\alpha C_i+(1-\alpha)D_i\right)}{\alpha (1-\alpha)}},
\end{align}
where $C_i=\KLr(\widehat{P}_{W,Z_i}\| P_W\otimes\mu)$ and $D_i=\KLr(\widehat{P}_{W,Z_i} \| P_{W,Z_i})$\,.
Finally, we prove the claim using \eqref{Eq: Ex 3}. 
\end{IEEEproof}

\begin{IEEEproof}[Proof of Theorem~\ref{thm: Minimum upper Bound alpha Renyi}]
    Using Lemma~\ref{lemma: alpha renyi equality}, we have:
\begin{align*}
    &\min_{\widehat{P}_{W,Z_i}} \alpha \KLr(\widehat{P}_{W,Z_i} \| P_W\otimes\mu)+(1-\alpha)\KLr(\widehat{P}_{W,Z_i} \| P_{W,Z_i})=\\\nn
    & (1-\alpha) I_{\mrR}^\alpha(W;Z_i)\\&+ \min_{\widehat{P}_{W,Z_i}} \KLr\left(\widehat{P}_{W,Z_i}\|\frac{(P_{Z_i}\otimes P_W)^\alpha (P_{W,Z_i})^{(1-\alpha)}}{\int_{\mathcal{W}\times \mathcal{Z}}(\mrd P_{Z_i}\otimes \mrd P_W)^\alpha (\mrd P_{W,Z_i})^{(1-\alpha)} }\right).
\end{align*}
Now by considering the $d\widehat{P}_{W,Z_i}=\frac{(\mrd P_{Z_i}\otimes \mrd P_W)^\alpha (\mrd P_{W,Z_i})^{(1-\alpha)}}{\int_{\mathcal{W}\times \mathcal{Z}}(\mrd P_{Z_i}\otimes \mrd P_W)^\alpha (\mrd P_{W,Z_i})^{(1-\alpha)} }$, the KL term would be equal to zero. The final result holds by using Proposition~\ref{Prop: reverse upper bound sigma subgaussian}. 
\end{IEEEproof}

\begin{IEEEproof}[Proof of Lemma~\ref{lemma: alpha renyi equality}]

Our proof is based on \cite[Theorem~30]{van2014renyi}. For $0\leq \alpha \leq 1$, we have:
\small
\begin{align}
    &\alpha \KLr(\widehat{P}_{W,Z_i} \| P_W\otimes\mu)+(1-\alpha)\KLr(\widehat{P}_{W,Z_i} \| P_{W,Z_i})\\
    &=\int_{\mathcal{W}\times \mathcal{Z}} d\widehat{P}_{W,Z_i}\log(d\widehat{P}_{W,Z_i})\\\nn&\quad-\int_{\mathcal{W}\times \mathcal{Z}} \widehat{P}_{W,Z_i}\log((d P_W\otimes \mrd P_{Z_i})^{\alpha}(\mrd P_{W,Z_i})^{(1-\alpha)})\\
    &=\int_{\mathcal{W}\times \mathcal{Z}} d\widehat{P}_{W,Z_i}\log(d\widehat{P}_{W,Z_i})\\\nn&\quad-\int_{\mathcal{W}\times \mathcal{Z}} d\widehat{P}_{W,Z_i}\log\left((\mrd P_W\otimes \mrd P_{Z_i})^{\alpha}(\mrd P_{W,Z_i})^{(1-\alpha)}\right)\\\nn
    &\quad+\log\left(\int_{\mathcal{W}\times \mathcal{Z}} (\mrd P_W\otimes \mrd P_{Z_i})^{\alpha}(\mrd P_{W,Z_i})^{(1-\alpha)}\right)\\\nn&\quad-\log\left(\int_{\mathcal{W}\times \mathcal{Z}} (\mrd P_W\otimes \mrd P_{Z_i})^{\alpha}(\mrd P_{W,Z_i})^{(1-\alpha)}\right)\\
    &=-\log\left(\int_{\mathcal{W}\times \mathcal{Z}} (P_W\otimes \mrd P_{Z_i})^{\alpha}(\mrd P_{W,Z_i})^{(1-\alpha)}\right)\\\nn&\quad
    +\int_{\mathcal{W}\times \mathcal{Z}} d\widehat{P}_{W,Z_i}\log(d\widehat{P}_{W,Z_i})
    \\\nn&\quad-\int_{\mathcal{W}\times \mathcal{Z}} d\widehat{P}_{W,Z_i}\log\left(\frac{(\mrd P_W\otimes \mrd P_{Z_i})^{\alpha}(\mrd P_{W,Z_i})^{(1-\alpha)}}{\int_{\mathcal{W}\times \mathcal{Z}} (\mrd P_W\otimes \mrd P_{Z_i})^{\alpha}(\mrd P_{W,Z_i})^{(1-\alpha)}}\right)\\
    &=(1-\alpha) I_{\mrR}^\alpha(W;Z_i)\\\nn&\quad+\KLr\left(\widehat{P}_{W,Z_i}\|\frac{(P_{Z_i}\otimes P_W)^\alpha (P_{W,Z_i})^{(1-\alpha)}}{\int_{\mathcal{W}\times \mathcal{Z}}(\mrd P_{Z_i}\otimes \mrd P_W)^\alpha (\mrd P_{W,Z_i})^{(1-\alpha)} }\right).
\end{align}
\normalsize
\end{IEEEproof}

\begin{IEEEproof}[Proof of Proposition~\ref{Proposition: decay rate renyi}]
Using \eqref{Eq: Renyi compare to KL},
\begin{align}
    I_{\mrR}^\alpha(W;Z_i)\leq \frac{\alpha}{1-\alpha} I(W;Z_i),
\end{align}
and considering the hypothesis space is finite and the upper bound in Theorem~\ref{thm: Minimum upper Bound alpha Renyi}, we have:
\begin{align}
    &|\genb(P_{W|S},\mu)|\leq \frac{1}{n}  \sum_{i=1}^n \sqrt{2(\alpha\sigma^2+(1-\alpha)\gamma^2)\frac{I_{\mrR}^\alpha(W;Z_i)}{\alpha}}\\
    &\leq \frac{1}{n}  \sum_{i=1}^n \sqrt{2(\alpha\sigma^2+(1-\alpha)\gamma^2) \min\left\{\frac{1}{\alpha},\frac{1}{1-\alpha}\right\}I(W;Z_i) }\\\label{eq: concavity}
    &\leq   \sqrt{2(\alpha\sigma^2+(1-\alpha)\gamma^2) \min\left\{\frac{1}{\alpha},\frac{1}{1-\alpha}\right\}\frac{\sum_{i=1}^n I(W;Z_i)}{n}}\\\label{eq: iid assumption}
    &\leq  \sqrt{2(\alpha\sigma^2+(1-\alpha)\gamma^2) \min\left\{\frac{1}{\alpha},\frac{1}{1-\alpha}\right\}\frac{I(W;S)}{n}}\\
    &\leq  \sqrt{2(\alpha\sigma^2+(1-\alpha)\gamma^2) \min\left\{\frac{1}{\alpha},\frac{1}{1-\alpha}\right\}\frac{H(W)}{n}}
      \\\nn&\quad\leq\sqrt{2(\alpha\sigma^2+(1-\alpha)\gamma^2) \min\left\{\frac{1}{\alpha},\frac{1}{1-\alpha}\right\}\frac{\log(k)}{n}},
\end{align}
where \eqref{eq: concavity} follows from Jensen inequality and \eqref{eq: iid assumption} follows from i.i.d assumption for $Z_i$'s.
\end{IEEEproof}
\begin{IEEEproof}[Proof of Theorem~\ref{Theorem: sibson result}]
Consider arbitrary auxiliary distributions $\{\tilde{P}_{W,Z_i}\}_{i=1}^n$ defined on $\mathcal{W}\times \mathcal{Z}$.
\begin{align}\label{Eq: Ex 33}
    &\genb(P_{W|S},\mu)=\mbE_{P_{W}P_{S}}[L_E(W,S)]-\mbE_{P_{W,S}}[L_E(W,S)]\nn\\
    &=\frac{1}{n}\sum_{i=1}^n \mbE_{P_{W}P_{Z_i}}[\ell(W,Z_i)]- \mbE_{P_{W Z_i}}[\ell(W,Z_i)] \\
    &\leq \frac{1}{n}\sum_{i=1}^n \left|\mbE_{P_{W}P_{Z_i}}[\ell(W,Z_i)]- \mbE_{P_{W Z_i}}[\ell(W,Z_i)]\right|.
\end{align}

Using the assumption centered loss function $\ell(w,z_i)-\mbE_{P_{Z_i}}[\ell(w,Z_i)]$ is $\gamma^2$-sub-Gaussian under distribution $P_{W,Z_i}$ and Donsker-Varadhan representation by considering function $\ell(w,z_i)-\mbE_{P_{Z_i}}[\ell(w,Z_i)]$ we have:
\begin{align}\label{p1001}
&\lambda\bigg(\mbE_{\tilde{P}_{W,Z_i}}[\ell(W,Z_i)-\mbE_{P_{Z_i}}[\ell(W,Z_i)]]\\\nn&\quad- \mbE_{P_{W Z_i}}[\ell(W,Z_i)-\mbE_{P_{Z_i}}[\ell(W,Z_i)]]\bigg ) \\\nn&\le \KLr(\tilde{P}_{W,Z_i}\|P_{W Z_i})+\frac{\lambda^2\gamma^{2}}{2}.\quad\forall\lambda\in \mathbb{R}
\end{align}
Note that $\mbE_{P_{W Z_i}}[\ell(W,Z_i)-\mbE_{P_{Z_i}}[\ell(W,Z_i)]]=\mbE_{P_{W Z_i}}[\ell(W,Z_i)]-\mbE_{P_W P_{Z_i}}[\ell(W,Z_i)]$.

Using the assumption that $\ell(w,Z)$ is $\sigma^{2}$-sub-Gaussian under $P_{Z_i}$ for all $w\in \mathcal{W}$, and again Donsker-Varadhan representation by considering function $\ell(w,z_i)-\mbE_{P_{Z_i}}[\ell(w,Z_i)]$ we have:
\begin{align}\label{p1002}
 &\lambda'\bigg( \mbE_{\tilde{P}_{W,Z_i}}[\ell(W,Z_i)-\mbE_{P_{Z_i}}[\ell(W,Z_i)]]\\\nn&\quad-\mbE_{Q_{W}P_{Z_i}}[\ell(W,Z_i)-\mbE_{P_{Z_i}}[\ell(W,Z_i)]]\bigg)\\\nn&\quad\le \KLr(\tilde{P}_{W,Z_i}\|Q_{W}P_{Z_i})+\frac{{\lambda'}^2\sigma^{2}}{2}.\quad\forall\lambda'\in \mathbb{R}
\end{align}
Note that $\mbE_{Q_{W}P_{Z_i}}[\ell(W,Z_i)-\mbE_{P_{Z_i}}[\ell(W,Z_i)]]=0$.

Now if we consider $\lambda>0$, then we choose $\lambda'=\frac{\alpha}{\alpha-1} \lambda$. Hence we have

\begin{align}\label{r1001}
&\mbE_{\tilde{P}_{W,Z_i}}[\ell(W,Z_i)-\mbE_{P_{Z_i}}[\ell(W,Z_i)]]\\\nn&\quad- \mbE_{P_{W Z_i}}[\ell(W,Z_i)-\mbE_{P_{Z_i}}[\ell(W,Z_i)]]  \\\nn&\quad  \le \frac{\KLr(\tilde{P}_{W,Z_i}\|P_{W Z_i})}{\lambda}+\frac{\lambda\gamma^{2}}{2},\quad\forall\lambda\in \mathbb{R^+}.
\end{align}
Using the assumption $\ell(w,Z)$ is $\sigma^{2}$-sub-Gaussian and again Donsker-Varadhan representation,
\begin{align}\label{r1002}
&- \mbE_{\tilde{P}_{W,Z_i}}[\ell(W,Z_i)-\mbE_{P_{Z_i}}[\ell(W,Z_i)]]\le \\\nn&\quad\frac{\KLr(\tilde{P}_{W,Z_i}\|Q_{W}P_{Z_i})}{|\lambda'|}+\frac{|\lambda'|\sigma^{2}}{2},\quad\forall\lambda'\in \mathbb{R^-}.
\end{align}

Now sum up the two Inequalities \eqref{r1001} and \eqref{r1002} to obtain,
\begin{align}\label{Eq: Ex 55}
    &\mbE_{P_W P_{Z_i}}[\ell(W,Z_i)]-\mbE_{P_{W Z_i}}[\ell(W,Z_i)]\le\\\nn
    &\frac{\alpha \KLr(\tilde{P}_{W,Z_i}\|P_{W,Z_i})+(1-\alpha)\KLr(\tilde{P}_{W,Z_i}\|Q_{W}P_{Z_i})}{\alpha\lambda}+\\\nn&\quad
    \frac{\lambda\gamma^{2}}{2}+\frac{\lambda\frac{\alpha}{1-\alpha}\sigma^{2}}{2},\quad\forall\lambda\in\mathbb{R^+}.
\end{align}
Taking infimum on $\tilde{P}_{W,Z_i}$ and using \cite[Theorem 30]{van2014renyi} that states
\[
(1-\alpha)\ren(P_1\|P_2)=\inf_{R}\{\alpha \KLr(R\|P_1)+(1-\alpha)\KLr(R\|P_2)\}
\]
Now, we have:
\begin{align}
    &(1-\alpha)\ren(P_{W,Z_i}\|Q_{W}P_{Z_i})=
\\\nn&{\inf_{\tilde{P}_{W,Z_i}}\{\alpha \KLr(\tilde{P}_{W,Z_i}\|P_{W,Z_i})+(1-\alpha)\KLr(\tilde{P}_{W,Z_i}\|Q_{W}P_{Z_i})\}}
\end{align}
and taking infimum on $Q_{W}$, we have:
\begin{align}\label{Eq: Ex 44}
   \inf_{Q_{W}} \ren(P_{W,Z_i}\|Q_{W}P_{Z_i})=I_{\mathrm{S}}^{\alpha}(Z_i;W).
\end{align}
Using \eqref{Eq: Ex 44} in \eqref{Eq: Ex 55}, we get:
\begin{align}\label{Eq: Ex 1}
    &\mbE_{P_W P_{Z_i}}[\ell(W,Z_i)]-\mbE_{P_{W Z_i}}[\ell(W,Z_i)]\le\\\nn
    &\quad \frac{(1-\alpha)I^{s}_{\alpha}(Z_i;W)}{\lambda\alpha} +\frac{\lambda\gamma^{2}}{2}+\frac{\lambda\frac{\alpha}{1-\alpha}\sigma^{2}}{2}\quad\forall\lambda\in\mathbb{R^+}.
\end{align}
Using the same approach for $\lambda\in R^-$, we have:
\begin{align}\label{Eq: Ex 2}
    &\mbE_{P_{W Z_i}}[\ell(W,Z_i)]-\mbE_{P_W P_{Z_i}}[\ell(W,Z_i)]\le\\\nn
    &\quad \frac{(1-\alpha)I^{s}_{\alpha}(Z_i;W)}{|\lambda|\alpha} +\frac{|\lambda|\gamma^{2}}{2}+\frac{|\lambda|\frac{\alpha}{1-\alpha}\sigma^{2}}{2},\quad\forall\lambda\in\mathbb{R^-}.
\end{align}
Considering \eqref{Eq: Ex 1} and \eqref{Eq: Ex 2}, we have a non-negative parabola in $\lambda$, whose discriminant must be non-positive, and we have:
\begin{align}
  &\left | \mbE_{P_W P_{Z_i}}[\ell(W,Z_i)]-\mbE_{P_{W Z_i}}[\ell(W,Z_i)]\right|\le\\\nn&\quad \sqrt{2(\alpha\sigma^2 + (1-\alpha)\gamma^2)\frac{I_{\mathrm{S}}^{\alpha}(Z_i;W)}{\alpha}}.
\end{align}
We prove the claim using \eqref{Eq: Ex 3}.

\end{IEEEproof}
\begin{IEEEproof}[Proof of Proposition~\ref{prop: gen_renyi_pinsker's charac}]
The Generalized Pinsker's inequality is introduced in \cite{van2014renyi}, as follows,
\begin{align}
    \mathbb{TV}(P,Q)^{2}\le \frac{2}{\alpha}\ren(P\|Q),\quad\alpha\in(0,1],
\end{align}
where $\mathbb{TV}(P,Q)=\int_{\mathcal{X}}|P(\mathrm{d}x)-Q(\mathrm{d}x)|$.
Denote $f:\mathcal{X}\to\mathbb{R}$ a bounded function $|f|\le L$, then
\begin{align}
    &\mbE_{P}[f(X)]-\mbE_{Q}[f(X)]=\\\nn&\quad
    \int f(x)(P(dx)-Q(dx))\le\\\nn&\quad \sup_{x}f(x)\cdot \int |P(dx)-Q(dx)|\le L \sqrt{\frac{2}{\alpha}\ren(P\|Q)}.
\end{align}
Let $P=P_{W,Z}$, $Q=P_{W}P_{Z}$ and $f(w,z)=L_{\mu}(w)-L_{E}(w,z)$. Then, we have the final result,
\begin{align*}
    &\genb(P_{W|S},\mu)=\frac{1}{n}\sum_{i=1}^{n}\mbE[L_{\mu}(W)-L_{E}(W,Z_{i})]\\&\le\frac{1}{n}\sum_{i=1}^{n}\sqrt{\frac{2b^{2}}{\alpha}\ren(P_{W,Z_{i}}\|P_{W}P_{Z_{i}})}.
\end{align*}
\end{IEEEproof}
\section{ Proof of Section~\ref{section: compare}}\label{app: sec compare}
\begin{IEEEproof}[Proof of Proposition~\ref{Proposition: alpha renyi to general Jensen-Shannon}]
    It follows from $$I_{\JDS}^\alpha(W;Z_i)\leq \frac{h(\alpha')}{\alpha' (1-\alpha')}, $$ that if we have $\frac{\alpha h(\alpha')}{\alpha' (1-\alpha')} \leq I_{\mrR}^\alpha(W;Z_i)$ for all $i=1,\cdots,n$, then the results holds for $\sigma=\gamma=\sigma_{JS}$.
\end{IEEEproof} 
\section{Proofs of Section~\ref{sec: excess risk}}
\begin{IEEEproof}[{Proof of Theorem~\ref{thm: upper true with JSD}}]
Let us define $P_{\mathcal{N}}:=\mathcal{N}(w^\star,\beta^{-1} I_d)$ and $w^\star:=\arg\inf_{w\in\mathcal{W}}L_{\mu}(w)$.
   \begin{align*}
        &\mathcal{E}_r(P_{W|S}^{\star,\beta,\JDS_{\alpha}},\mu) \\& \leq 
        \big|\genb(P_{W|S}^{\star,\beta,\JDS_{\alpha}},\mu)\big|+ \mbE_{P_S\otimes P_{W|S}^{\star,\beta,\JDS_{\alpha}}}[L_{E}(W,S)]-L_{\mu}(w^\star)\\
        &\leq  \sqrt{\frac{2b^2}{n\alpha(1-\alpha)}\sum_{i=1}^n I_{\JDS}^\alpha(W;Z_i)}+ \mbE_{P_S\otimes P_{\mathcal{N}}}[L_{E}(W,S)]-L_{\mu}(w^\star)\\&\quad+\frac{\JDS_{\alpha}(\mathcal{N}(w^\star,\beta^{-1} I_d)\|Q)}{\beta}\\
        &\leq  \sqrt{\frac{2b^2}{n\alpha(1-\alpha)}\sum_{i=1}^n I_{\JDS}^\alpha(W;Z_i)}+ \mbE_{P_{\mathcal{N}}}[L_{\mu}(W)]-L_{\mu}(w^\star)\\&\quad+\frac{\JDS_{\alpha}(\mathcal{N}(w^\star,\beta^{-1} I_d)\|Q)}{\beta}\\
         &\leq  \sqrt{\frac{2b^2}{n\alpha(1-\alpha)}\sum_{i=1}^n I_{\JDS}^\alpha(W;Z_i)}+ \mbE_{P_{\mathcal{N}}}[L_{\mu}(w^\star)\\&\quad+\tilde{L}\|W-w^\star\|_2]-L_{\mu}(w^\star)+\frac{\JDS_{\alpha}(\mathcal{N}(w^\star,\beta^{-1} I_d)\|Q)}{\beta}\\
        &\leq \sqrt{\frac{2b^2}{n\alpha(1-\alpha)}\sum_{i=1}^n I_{\JDS}^\alpha(W;Z_i)}+\frac{\tilde{L}\sqrt{d}}{\beta}\\&\quad+\frac{\JDS_{\alpha}(\mathcal{N}(w^\star,\beta^{-1} I_d)\|Q)}{\beta},
    \end{align*}
Note that $\tilde{L}\mbE_{P_{\mathcal{N}}}[\|W-w^\star\|_2]=\frac{\tilde{L}d}{\beta}$.
\end{IEEEproof}

\begin{IEEEproof}[Proof of Theorem~\ref{thm: upper true with alpha renyi}]
    The proof is similar to Proof of Theorem~\ref{thm: upper true with JSD}, by replacing the second inequality with the following inequality,
    \begin{equation*}
        \begin{split}
             &\mathcal{E}_r(P_{W|S}^{\star,\beta,\mrR_{\alpha}},\mu)  \leq 
        \big|\genb(P_{W|S}^{\star,\beta,\mrR_{\alpha}},\mu)\big|\\&\quad+ \mbE_{P_S\otimes P_{W|S}^{\star,\beta,\JDS_{\alpha}}}[L_{E}(W,S)]-L_{\mu}(w^\star)\\
        &\leq  \sqrt{\frac{2b^2}{n\alpha(1-\alpha)}\sum_{i=1}^n I_{\mrR}^\alpha(W;Z_i)}\\&\quad+ \mbE_{P_S\otimes P_{\mathcal{N}}}[L_{E}(W,S)]-L_{\mu}(w^\star)+\frac{\ren(\mathcal{N}(w^\star,\beta^{-1} I_d)\|Q)}{\beta}
        \end{split}
    \end{equation*}
\end{IEEEproof}

\begin{IEEEproof}[Proof of Corollary~\ref{cor: conv rate of excess risk jsd}]
     Using the boundedness of $\alpha$-$\JDS$ divergence in Theorem~\ref{thm: upper true with JSD}, we have,
    \[
    \begin{split}
        \mathcal{E}_r(P_{W|S}^{\star,\beta,\JDS_{\alpha}},\mu)&\leq  \sqrt{\frac{2b^2}{n\alpha}\sum_{i=1}^n I_{\mrR}^\alpha(W;Z_i)}+\frac{\tilde{L}\sqrt{d}}{\beta}+\frac{h(\alpha)}{\beta},
    \end{split}
    \]
    where $h(\alpha)=-\alpha\log(\alpha)-(1-\alpha)\log(1-\alpha)$. Therefore, by setting $\beta=n^{1/2}$, the convergence rate of excess risk is $\mathcal{O}(1/\sqrt{n})$.
\end{IEEEproof}

\begin{IEEEproof}[Proof of Corollary~\ref{cor: conv rate renyi excess}]

 We consider the normal distribution as prior, i.e., $Q=\mathcal{N}(0,I_d)$, in Theorem~\ref{thm: upper true with alpha renyi}. Then, we can compute the $\alpha$-R\'enyi divergence between two multivariate Gaussian distributions~\cite{gil2013renyi},
 \[\begin{split}
     &\ren(\mathcal{N}(w^\star,\beta^{-1} I_d)\|\mathcal{N}(0,I_d))=\frac{\alpha}{2}\|w^\star\|_2^2(\alpha+(1-\alpha)\beta^{-1})^{-1}\\&\quad+\frac{d}{2(\alpha-1)}\log\Big( \frac{\beta^{\alpha-1}}{\alpha+(1-\alpha)\beta^{-1}}\Big).
     \end{split}\]
 Then, the following upper bound holds on the excess risk under $P_{W|S}^{\star,\beta,\mrR_{\alpha}}$,
 \[
    \begin{split}
         &\mathcal{E}_r(P_{W|S}^{\star,\beta,\mrR_{\alpha}},\mu)\leq  \sqrt{\frac{2b^2}{n\alpha}\sum_{i=1}^n I_{\mrR}^\alpha(W;Z_i)}+\frac{\tilde{L}\sqrt{d}}{\beta}\\&\quad+\frac{\alpha}{2\beta}\|w^\star\|_2^2(\alpha+(1-\alpha)\beta^{-1})^{-1}\\&\quad+\frac{d}{2\beta(\alpha-1)}\log\Big(\frac{\beta^{(\alpha-1)}}{\alpha+(1-\alpha)\beta^{-1}}\Big)\\
         &\leq \sqrt{\frac{2b^2}{n\alpha}\sum_{i=1}^n I_{\mrR}^\alpha(W;Z_i)}+\frac{\tilde{L}\sqrt{d}}{\beta}\\&\quad+\frac{1}{2\beta}\|w^\star\|_2^2+\frac{d}{2\beta}\log\big(\beta\big)+\frac{d}{2\beta(1-\alpha)}\log\big(\alpha\big).
    \end{split}
    \]
\end{IEEEproof}

 \section{Proofs of Section~\ref{Sec: mismatch}}\label{App: Proof of mismatch}
 We first propose the following Lemma to provide an upper bound on the expected generalization error under distribution mismatch.
 \begin{lemma}\label{Lemma: mismatch}
 The following upper bound holds on expected generalization error under distribution mismatch between the test and training distributions:
 \begin{align}\label{eq;mismatch}
     &|\genb(P_{W|S},\mu,\mu')|\leq \\\nonumber&\quad|\genb(P_{W|S},\mu)|+ |\mathbb{E}_{P_W\otimes \mu'}[\ell(W,Z)]-\mathbb{E}_{P_W\otimes \mu}[\ell(W,Z)]|
     .
 \end{align}
 \end{lemma}
 \begin{IEEEproof}
  We have:
 \begin{align}
&|\genb(P_{W|S},\mu,\mu')|\\\nonumber&
=|\mathbb{E}_{P_{W,S}}[L_P(W,\mu')-L_P(W,\mu)+L_P(W,\mu)-L_E(E,S)]|\\\nonumber&\le |\mathbb{E}_{P_{W,S}}[L_P(W,\mu')-L_P(W,\mu)]|+|\genb(P_{W|S},\mu)|\\\nonumber
&=|\mathbb{E}_{P_W\otimes \mu'}[\ell(W,Z)]-\mathbb{E}_{P_W\otimes \mu}[\ell(W,Z)]|+|\genb(P_{W|S},\mu)|
 \end{align}
\end{IEEEproof}
\begin{IEEEproof}[Proof of Proposition~\ref{Proposition: alpha general js upper Bound under mismatch}]
 In Lemma~\ref{Lemma: mismatch}, the generalization error under distribution mismatch can be upper bounded by two terms.
 Considering Theorem~\ref{thm: Minimum general js upper Bound}, we can provide the upper bound based on  $\alpha$-Jensen-Shannon information over $|\genb(P_{W|S},\mu)|$. We can also provide an upper bound on the term $|\mathbb{E}_{P_W\otimes \mu'}[\ell(W,Z)]-\mathbb{E}_{P_W\otimes \mu}[\ell(W,Z)]|$ in Lemma~\ref{Lemma: mismatch} by applying ADM using a similar approach as in Theorem~\ref{thm: Minimum general js upper Bound} and using the  $\alpha$-Jensen-Shannon divergence as follows:
 \begin{align}
     &|\mathbb{E}_{P_{W}\otimes \mu'}[\ell(W,Z)]-\mathbb{E}_{P_{W}\otimes \mu}[\ell(W,Z)]|\leq\\
     &\sqrt{2\sigma_{(\alpha)}^2\frac{\JDS_\alpha (P_W \otimes \mu'\|P_W \otimes \mu)}{\alpha (1-\alpha)}}=\\
     &\sqrt{2\sigma_{(\alpha)}^2\frac{\JDS_\alpha (\mu'\|\mu)}{\alpha (1-\alpha)}}.
 \end{align}
 \end{IEEEproof}
 \begin{IEEEproof}[Proof of Proposition~\ref{Prop: upper Bound alpha Renyi under mismatch}]
 Based on Lemma~\ref{Lemma: mismatch}, the generalization error is upper bounded by two terms (See Equation \eqref{eq;mismatch}). We can provide the upper bound based on $\alpha$-R\'enyi information over $|\genb(P_{W|S},\mu)|$ using Theorem~\ref{thm: Minimum upper Bound alpha Renyi}.  We can also provide an upper bound on the term $|\mathbb{E}_{P_W\otimes \mu'}[\ell(W,Z)]-\mathbb{E}_{P_W\otimes \mu}[\ell(W,Z)]|$ by applying ADM using a similar approach as in Theorem~\ref{thm: Minimum upper Bound alpha Renyi} and using $\alpha$-R\'enyi divergence as follows:
 \begin{align}
     &|\mathbb{E}_{P_{W}\otimes \mu'}[\ell(W,Z)]-\mathbb{E}_{P_{W}\otimes \mu}[\ell(W,Z)]|\leq\\
     &\sqrt{2(\alpha\sigma^2+(1-\alpha)\gamma^2)\frac{\ren(P_W \otimes \mu'\| P_W \otimes \mu)}{\alpha}}=\\
     &\sqrt{2(\alpha\sigma^2+(1-\alpha)\gamma^2)\frac{\ren(\mu'\| \mu)}{\alpha}}.
 \end{align}
 \end{IEEEproof}

\end{document}